\documentclass[journal]{IEEEtran}
\usepackage{url}
\usepackage{cite}
\usepackage{rotating}
\usepackage{multirow}
\usepackage{multicol}
\usepackage{graphicx}
\usepackage{booktabs}
\usepackage{bbding}
\usepackage{pifont}
\usepackage{color}
\usepackage{relsize}
\usepackage{comment}
\usepackage{textcomp}
\usepackage{mathrsfs}
\usepackage{amsfonts}
\usepackage{amsmath} 
\usepackage{amssymb}  
\usepackage{multicol}
\usepackage{subeqnarray}
\usepackage{cases}
\usepackage{bm}
\usepackage{subfigure}
\usepackage{threeparttable}

\begin{document}

\title{Survey of Deep Learning for Autonomous Surface Vehicles in Marine Environments}

\author{Yuanyuan~Qiao,~\IEEEmembership{Member,~IEEE,}
        Jiaxin~Yin,
        ~Wei~Wang,~\IEEEmembership{Member,~IEEE,}
        ~Fábio~Duarte, 
        ~Jie~Yang,
        ~and~Carlo~Ratti
\thanks{ \textit{(Corresponding authors: Yuanyuan Qiao, and Wei Wang.)}}
\thanks{Yuanyuan Qiao is with Intelligent Perception and Computing Research Center, the School of Artificial Intelligence, Beijing University of Posts and Telecommunications (BUPT), Beijing 100876, China, and also with the MIT Senseable City Lab, Cambridge, MA 02139 USA (e-mail: yyqiao@bupt.edu.cn).}
\thanks{Jiaxin Yin, and Jie Yang are with Intelligent Perception and Computing Research Center, the School of Artificial Intelligence, Beijing University of Posts and Telecommunications (BUPT), Beijing 100876, China (e-mail: yinjx@bupt.edu.cn; janeyang@bupt.edu.cn).}
\thanks{Wei Wang, Fábio Duarte, and Carlo Ratti are with the MIT Senseable City Lab, Cambridge, MA 02139 USA (e-mail: wweiwang@mit.edu; fduarte@mit.edu; ratti@mit.edu).}
}

\maketitle
\IEEEpeerreviewmaketitle

\begin{abstract}
Within the next several years, there will be a high level of autonomous technology that will be available for widespread use, which will reduce labor costs, increase safety, save energy, enable difficult unmanned tasks in harsh environments, and eliminate human error. Compared to software development for other autonomous vehicles, maritime software development, especially in aging but still functional fleets, is described as being in a very early and emerging phase. This presents great challenges and opportunities for researchers and engineers to develop maritime autonomous systems. Recent progress in sensor and communication technology has introduced the use of autonomous surface vehicles (ASVs) in applications such as coastline surveillance, oceanographic observation, multi-vehicle cooperation, and search and rescue missions. Advanced artificial intelligence technology, especially deep learning (DL) methods that conduct nonlinear mapping with self-learning representations, has brought the concept of full autonomy one step closer to reality. This article reviews existing work on the implementation of DL methods in fields related to ASV. First, the scope of this work is described after reviewing surveys on ASV developments and technologies, which draws attention to the research gap between DL and maritime operations. Then, DL-based navigation, guidance, control (NGC) systems and cooperative operations are presented. Finally, this survey is completed by highlighting current challenges and future research directions. 


\end{abstract}



\section{Introduction} 
\IEEEPARstart{I}{n} the next decades, water, air, and land transport will be deeply shaped by autonomous vehicles. 
Although many technological challenges have not yet been solved, autonomous systems will undoubtedly be the core component of future transportation systems \cite{ab2017emerging}. Since the last century, unmanned aerial vehicles (UAVs) and autonomous underwater vehicles (AUVs) have been widely deployed in a variety of real-world applications. 
As a result of the efforts of leading technology companies, autonomous cars have been tested for millions of miles in preparation for full commercialization. 
Having benefitted from the advancement of guidance and control theory for surface vehicles, 
for decades, ASVs have been deeply involved in military, research, and commercial applications, including surveillance, data collection, and sea, surface and space communication hubs \cite{liu2016unmanned}. 

ASV can minimize the impact, limitation, and cost of human operators. After being launched from a dock, an ASV can be remotely operated by a human. With the help of computers, global positioning systems (GPSs), differential global positioning systems (DGPSs), and satellite communications, ASVs can navigate and perform a task autonomously \cite{manley2008unmanned}. In autonomous mode, ASV can perform a given mission without external supervision and then return to the dock at the end of its mission. At the beginning of the 20$^{th}$ century, the lack of effective and reliable obstacle detection sensors slowed the emergence of reliable obstacle avoidance methods \cite{caccia2006autonomous}. Currently, the advent of more sophisticated airborne and satellite sensors has enabled the study of temperature, moisture, and wind fields in maritime convective systems. Advanced sensors, such as light detection and ranging (LiDAR), are characterized by strong computing capabilities and high-accuracy positioning systems with global coverage and have already reshaped the navigation system of all autonomous vehicles \cite{zolich2019survey}. New data transfer technologies, high-capacity local area networks (LANs), wide area networks (WANs), and inexpensive satellite-based data communications are expected to further international collaborations toward the development of ASVs \cite{stateczny2022wireless}.

Recently, the capability of ASVs was greatly improved by large-scale data with machine learning and artificial intelligence technologies \cite{thombre2021sensors,karimi2021guidance}. DL methods, such as deep neural networks (DNNs) and deep reinforcement learning (DRL), have been used to make progress in solving long-existing problems of ASVs that cannot be solved with traditional methods \cite{perera2020deep}. For example,
\begin{itemize}
\item \textbf{Nonlinear System Identification:} As a universal approximator,
DNN can estimate the unknown parameters of the ASV model \cite{tee2006control,zheng2017trajectory}, unknown dynamics \cite{moreira2003dynamic,shojaei2016observer} and environmental disturbances. In addition, using a learning function that dynamically maps the relationship between the input variable (state variable) and the output variable (hydrodynamic force and moment data), a controller can be built without \textit{a priori} knowledge of the ASV dynamics or a mathematical model \cite{woo2019deep}.

\item \textbf{Model-free Control:} In complicated maritime environments, it is almost impossible to model the exact dynamics of a nonlinear system \cite{wang2021reinforcement}. DL methods can be used to design optimal data-driven and model-free control that requires only the ASV input-output signals without knowing the model \cite{wang2021data}. Furthermore, DRL can characterize and control extremely complex systems learning the interaction between agent and environment \cite{arulkumaran2017deep,kiran2021deep,zhao2020path}.

\item \textbf{Sea Surface Object Detection:} Based on a DL method, a sufficient amount of relevant data is collected by the sensors, which allows an accurate estimate of the state of a vehicle. 
Environmental information, including other ASVs 
and the harbor \cite{jin2020patch}, can be detected and analyzed by applying a DL method to different types of data sources \cite{lecun2015deep}.

\item \textbf{Future Behavior Prediction:} By exploring long-term historical automatic identification system (AIS) data based on DL methods, the future status of maritime transportation systems is available for advanced management and planning \cite{wen2020automatic} toward safer, highly efficient, and energy-conscientious maritime systems \cite{xiao2019traffic}.

\item \textbf{Human Like Decision Making:} To make decisions while complying with regulations when encountering other surface vehicles, a DL-based method is able to learn possible human behaviors from complex task experiences such as collision avoidance \cite{woo2020collision} and docking \cite{shuai2019efficient}. DL methods can automatically learn high-level features to react appropriately in complex environments with many constraints \cite{mnih2015human}.
\end{itemize}

This work comprehensively summarizes and compares the application of DL methods to ASVs and how DL techniques have permeated the entire field. Related topics include, but are not limited to, NGC systems and cooperative operations, as well as the integration and application of advanced sensors, communication systems, and big data technologies. This article concludes by discussing challenges and presenting possible future research directions that may be worth pursuing based on DL methods. In general, the aims of this work are as follows. 
\begin{enumerate}[]
\item Present a current survey of research showing how DL techniques improve ASV systems and successfully solve emerging challenges. In particular, advances in DL-based NGC systems are highlighted. 
\item Further, we discuss current research and future directions for the application of DL techniques from the point of view of intelligent maritime operations.
\end{enumerate}

The authors hope that this work can guide researchers, engineers and managers who want to understand the application of DL for maritime operations or employ DL to solve related problems. First, in Section \ref{sec:existing_survey}, existing ASV-related surveys are reviewed and the scope of this work is defined. In Section \ref{sec:systems_prototypes}, an overview of the basic ASV structure is provided, including hardware, onboard equipment, and the NGC system. 
 DL applications for navigation systems (Section \ref{sec:navigation}), guidance system (Section \ref{sec:guidance}), and control system (Section \ref{sec:control}) are examined and compared in detail, with a focus on the implementation of gradually evolving DL models on NGC systems. Then, cooperative maritime operations are reviewed in Section \ref{sec:DL_fleet}. The research gaps, current challenges, and future research directions are discussed in Section \ref{sec:challenge}. Finally, the conclusion of this work is given in Section \ref{sec:conclusion}. 

\section{ASV-Related Surveys and Survey Scope}
\label{sec:existing_survey}
From 2006 to 2017, only 16 surveys were published, and they mainly covered ASV prototypes \cite{caccia2006autonomous, bertram2008unmanned, manley2008unmanned,rynne2009unmanned,yan2010development,campbell2012review,stelzer2011history,zheng2013survey,othman2015review,manley2016unmanned,liu2016unmanned,schiaretti2017survey2} and NGC systems \cite{rynne2009unmanned,ashrafiuon2010review,campbell2012review,azzeri2015review,liu2016unmanned}. At the beginning of 2017, many large ship companies, such as Rolls-Royce, Kongsberg, and Yara, made contributions to the development of an autonomous ship. Rolls-Royce published a report and made efforts toward its vision of autonomous shipping \cite{levander2017autonomous}. Then, Kongsberg and Yara collaborated to build the world’s first fully electric, autonomous, zero-emission ship - \textit{Yara Birkeland}. At the end of 2017, the International Maritime Organization (IMO) developed their Strategic Plan for the Organization for the 6-year period from 2018 to 2023 and discussed the issue of autonomous marine surface ships, which was defined as MASS in 2018.

Between 2018 and 2022, 25 surveys strongly related to ASVs were published. Some of these works introduced further developments of the ASV and NGC systems \cite{polvara2018obstacle,wang2019review,silva2019rigid,huang2020ship,jing2020path,zhou2020review,vagale2021path,thombre2021sensors,karimi2021guidance,gu2022advances}, as well as their real-world applications \cite{zereik2018challenges,moud2018current,verfuss2019review,jorge2019survey}. In addition, to meet the goal of introducing the next level of autonomy to ASVs, the researchers focused their attention on collaboration \cite{liu2018survey,thompson2019review,chen2020survey} and communication \cite{ge2018wireless} between multiple ASVs and other vehicles. Other researchers explored current trends toward autonomous shipping \cite{van2018survey,munim2019autonomous,wang2019state,huang2020ship,zhang2021collision}. When imagining ASV or MASS voyaging on a surface autonomously, issues, including control and support centers \cite{researchinmaritime}, civil liabilities and insurance \cite{coremaritime}, technology \cite{martin2019advancing} and business requirements \cite{munim2019autonomous}, need to be addressed carefully. 

DL has radically changed many research areas and led to a surge of research in the past decade. 
DNNs are capable of forming compact representations of states from raw, high-dimensional, multi-modal sensor data, which are commonly found in robotic systems. 
For navigation subsystems, convolutional neural networks (CNNs) with hierarchical feature extraction capability have already been shown to be efficient for object detection and obstacle identification \cite{jorge2019survey}. With the collection of increasing amounts of maritime data, DL is considered a powerful framework compared to other machine learning methods to interpret large quantities of data automatically and relatively quickly. 
DNN has also been shown to be efficient in ASV control, especially in autonomous docking \cite{wang2019state}. 
In the field of robotic control for intelligent autonomous systems, DRL \cite{arulkumaran2017deep} has recently been applied to AUV, 
UAV \cite{carrio2017review}, autonomous cars \cite{kiran2021deep}, and ASVs \cite{woo2019deep}. 

However, based on existing surveys, there is a lack of discussion about the implementation of DL for ASVs. Therefore, this work reviews past and future technical developments and challenges of ASVs, especially the role of DL methods in improving the intelligence and autonomous levels. In addition to a thorough comparison of ASV prototypes and the architecture of ASV control systems, the authors try to answer the following questions: ``\textit{What DL methods have been successfully applied to ASVs? What are the theoretical and practical strengths and weaknesses}"; ``\textit{How can current maritime operations be solved with DL methods? and, which directions are promising for future research?}" To the best knowledge of the authors, the DL techniques for ASVs have not yet been comprehensively studied. A better understanding of the current and potential role of up-and-coming DL techniques in ASVs will provide momentum for future autonomous maritime operations.

\section{ASV Systems}
\label{sec:systems_prototypes}


\subsection{ASV Category}
\begin{figure}
\centering 
\includegraphics[width=0.45\textwidth]{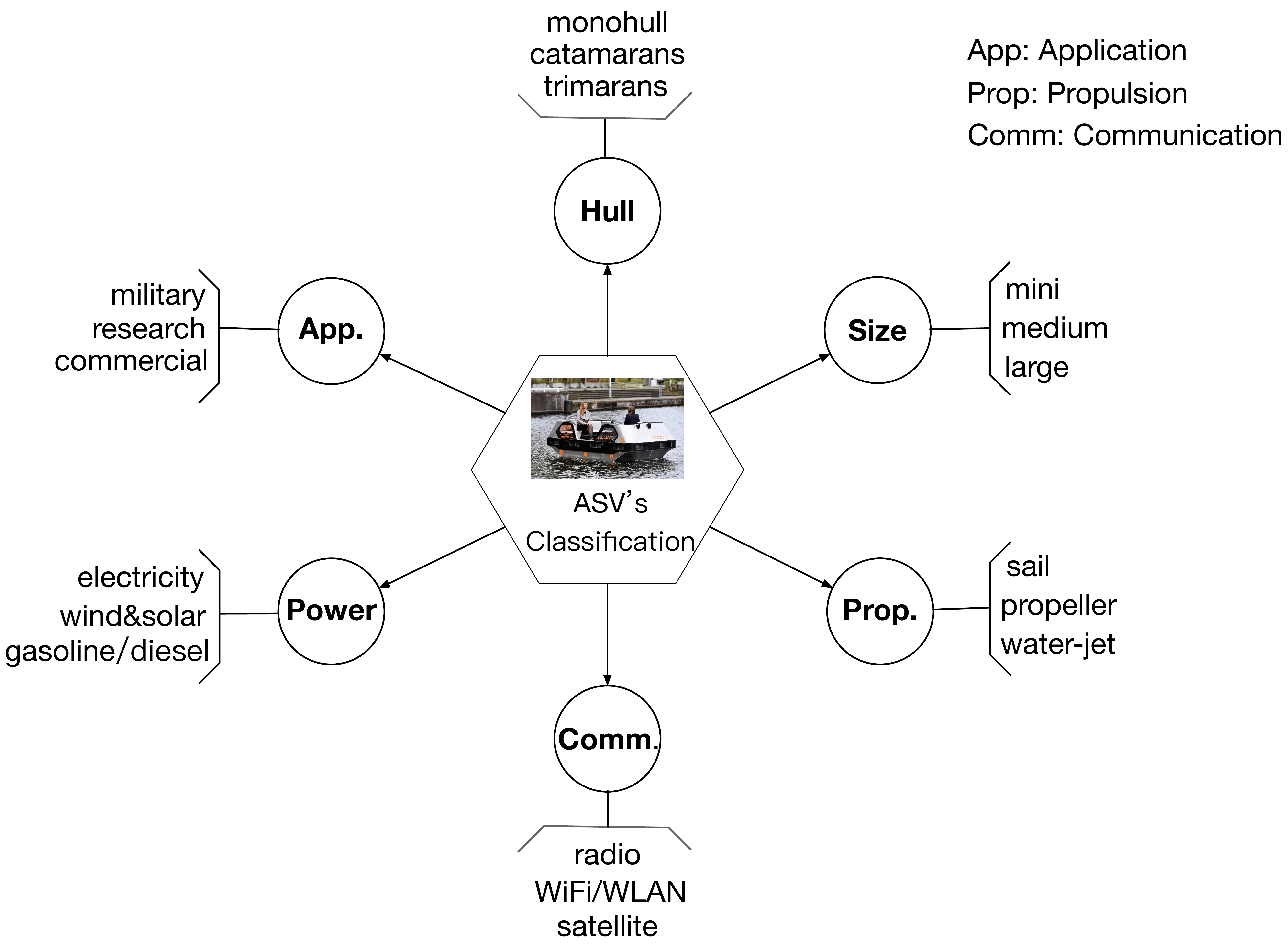} 
\caption{ASV Classification} 
\label{classification} 
\end{figure}

Each type of ASV can be leveraged for a specific application based on its size, function, or unique features, as shown in \figurename~\ref{classification}. For example, naval ASVs need to be controlled remotely and operated at high speed. Additionally, heavy payloads and limitations on energy consumption are required.
ASV research for oceanography requires autonomous navigation, energy conservation, and stable measurements at low speed \cite{manley2016unmanned}. Commercial surveillance ASVs are usually equipped with advanced visual monitoring sensors, high-performance controllers, and efficient communication systems with on-shore management centers.

\subsection{ASV Architecture}


\label{Ship motion}
\begin{figure}
\centering 
\includegraphics[width=0.33\textwidth]{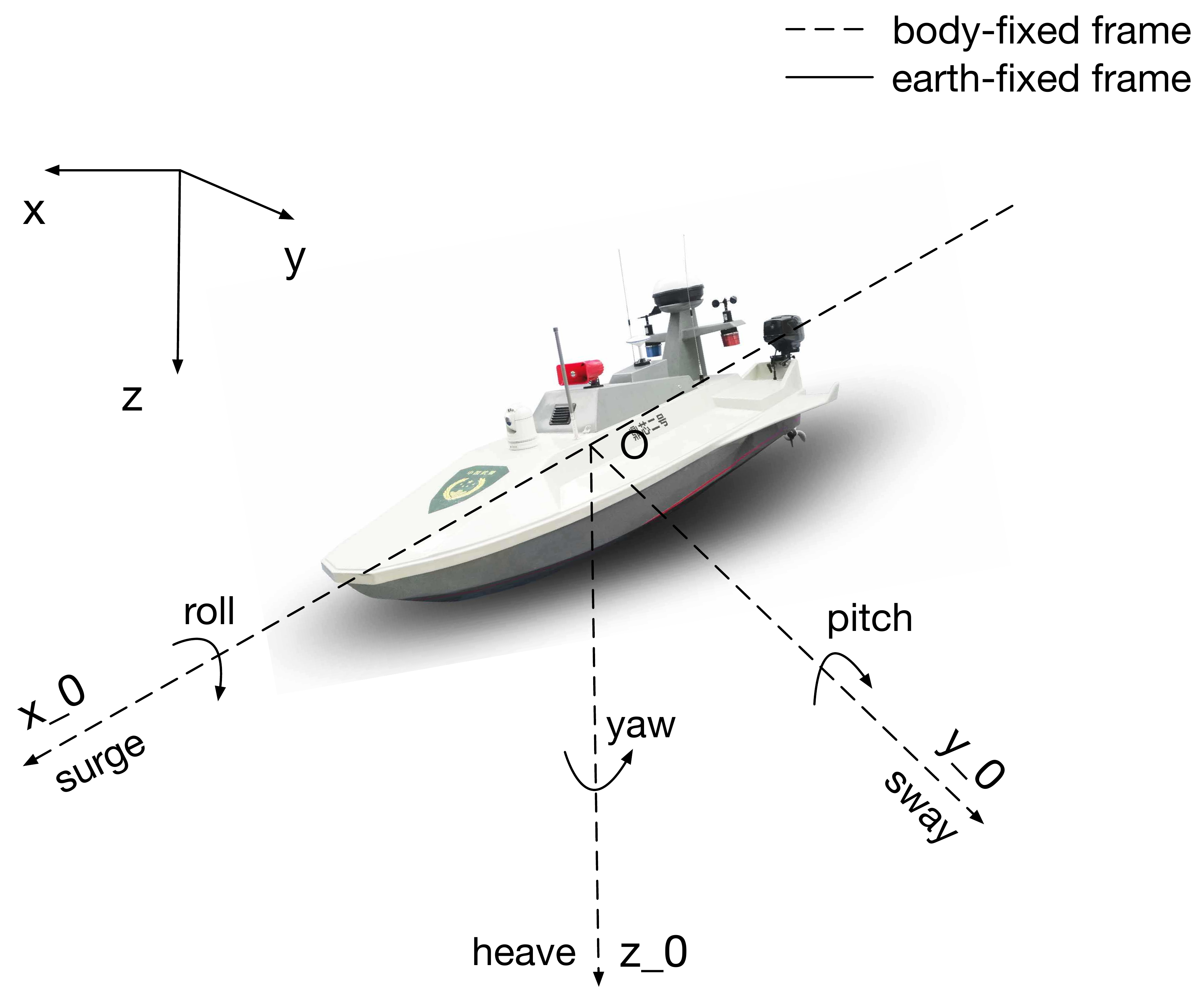} 
\caption{The 6 DOFs of an ASV} 
\label{ship_DOF} 
\end{figure}

The motion of a ship can be described as an earth-fixed frame or body-fixed frame, as shown in \figurename~\ref{ship_DOF}. The rigid body motions of a typical surface vehicle with 6 DOFs include:

(1) Three types of displacement motions (heave, sway or drift, and surge), which move through the x\_0, y\_0, and z\_0 directions, respectively. 

(2) Three angular motions (yaw, pitch, and roll), which are the rotations about the x\_0, y\_0, and z\_0 axes, respectively. 

Among the multiple options for the manufacture of ASV, the hull supports the main four components of the system, including the engine, communication, sensor and NGC systems \cite{schiaretti2017survey2}.

\begin{figure}
\flushright
\includegraphics[width=0.45\textwidth]{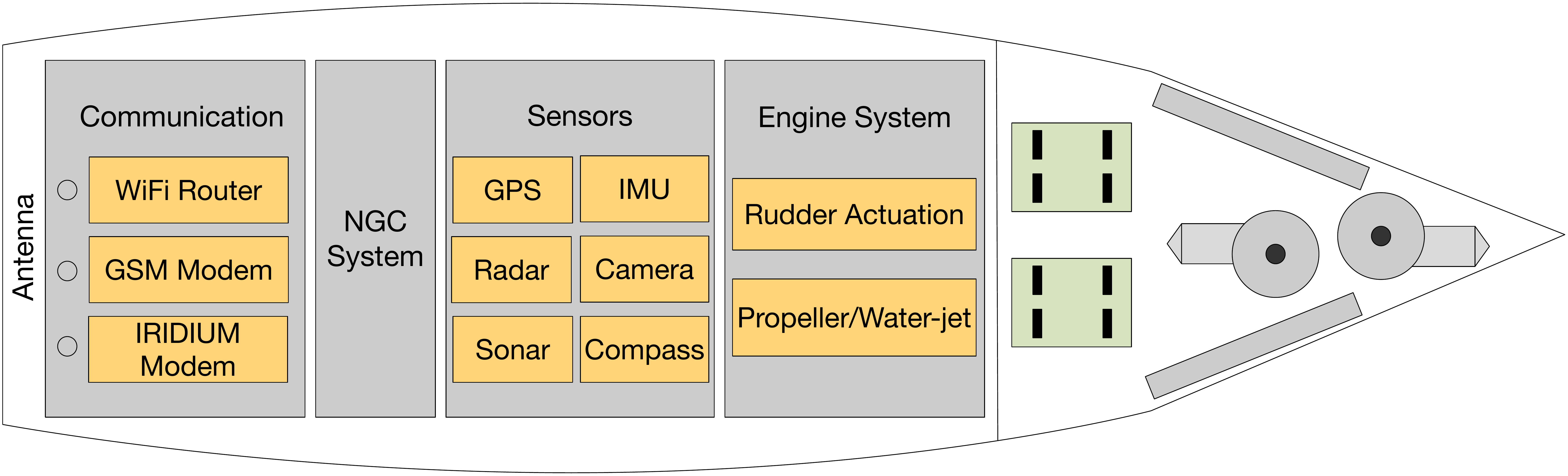} 
\caption{The Architecture of a typical ASV} 
\label{hardware} 
\end{figure}

\subsubsection{Hull}
As the main physical component of ASVs, hulls are waterproof bodies of ships or boats that can be classified into single hulls (kayaks and monohulls) and multihulls (catamarans and trimarans).

\subsubsection{Engine System}
The controller design should consider how many appreciable degrees of freedom (DOFs) of the ASVs can be actuated by the actuators. Most existing ASVs use underactuated controls, which means that only part of the DOFs can be controlled. If all DOFs can be controlled by multiple actuators, then the ASV is activated. For more complex tasks, such as docking, overactuated controls that can make use of additional DOFs are more effective.  

\subsubsection{Communication System}
Stable and reliable communication is essential for information exchange between (1) computers, sensors, and other hardware that need to be controlled; (2) multiple vehicles, such as ASVs, AUVs, and UAVs; (3) vehicles and onshore control centers; and (4) vehicles and remote satellites.

\subsubsection{Sensor System}
Sensors act as an interface between an ASV and the environment, providing a vehicle with information relative to its self-state and the environment. In addition to the performance monitoring sensor, the ASV shipboard sensors provide location, status, and environmental information, as shown in Table III. The deployment of different types of sensors depends on the mission requirements.

\subsubsection{NGC System}
\begin{figure}
\centering 
\includegraphics[width=0.4\textwidth]{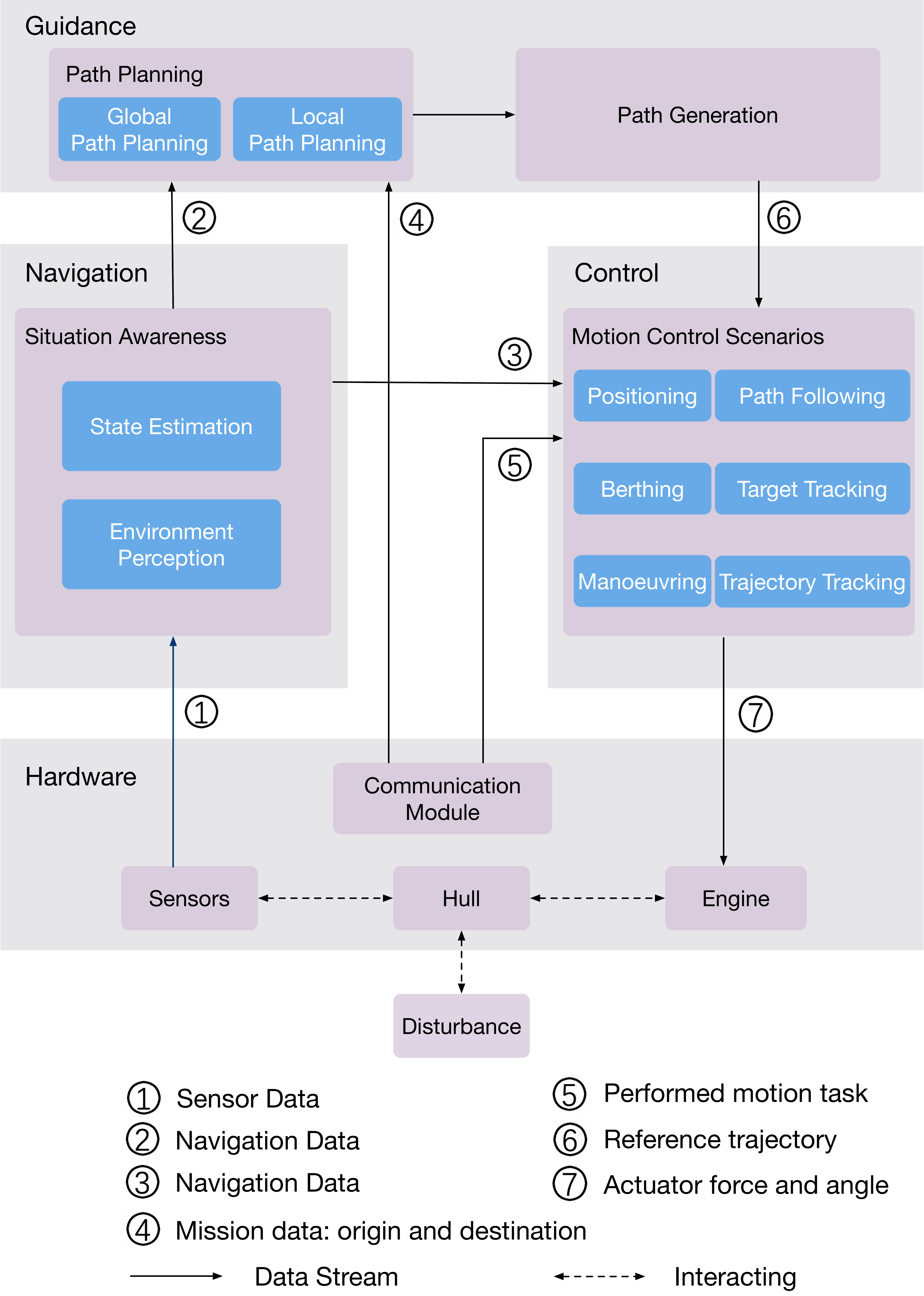} 
\caption{The structure of ASV systems} 
\label{ngc} 
\end{figure}

In an attempt to automate the ASV operation process, three systems are indispensable in the software structure. Navigation, guidance, and control, as illustrated in \figurename~\ref{ngc}. As software installed on an onboard computer, the NGC system processes the data collected for situation awareness, plans the possible path, and drives the surface vehicle to the destination. The NGC system is the core of autonomous operation. 

$\bullet$ \textbf{Navigation} (Section \ref{sec:navigation}): Navigation usually refers to the study of the process of monitoring the movement of vehicles. As the most fundamental requirement for safe operation, full assessment of surface vehicle status and its surrounding environment based on sensor data is essential for maritime vehicles. There are two stages in navigation systems, that is, environmental perception and state estimation.



\textit{Environmental Perception} refers to the process of measuring surrounding environments, such as waves, winds, objects, and obstacles on the surface \cite{elkins2010autonomous}. 

\textit{State Estimation} refers to the problem of reconstructing the underlying state of a system given a sequence of measurements and a prior model of the system. 

$\bullet$ \textbf{Guidance} (Section \ref{sec:guidance}): Based on the status information provided by the navigation system, the goal of guidance is to plan and generate possible paths from departure to destination with the ability to avoid obstacles and obey rules.

\textit{Path Planning} can be further divided into global and local methods \cite{liu2016unmanned}. Global path planning aims to find an optimal or near-optimal collision-free path between the origin and destination in advance.
Local path planning requires an ASV response to the previously unknown obstacle and changes in the environment. 

\textit{Path Generation} refers to the process of generating an optimal path for ASV. 

$\bullet$ \textbf{Control} (Section \ref{sec:control}): 
The control system generates the proper control forces and moments for the ASV driving equipment \cite{liu2016unmanned}.

\textit{Motion Control Scenarios} refer to the specific scenarios in which an ASV operates, including point stabilization, target tracking, path following, trajectory tracking, maneuvering, and berthing. The definition of each motion control scenario can be found in Section \ref{sec:control}.

\section{Deep Learning Models and Techniques}
\label{sec:Enable_DL}

Many publication have focused on the advancement of DL \cite{lecun2015deep,bengio2021deep}, as well as the applications of DL on robotics \cite{sunderhauf2018limits}, autonomous cars \cite{grigorescu2020survey}, UAVs \cite{carrio2017review}, and UUVs. 
Therefore, to avoid unnecessary repetition, we briefly introduce the DL models and its applications categorized by different learning methods, i.e., supervised, unsupervised, and reinforcement learning methods.

\subsection{Supervised Deep Learning Method}
Supervised deep learning methods are trained using well-labelled data to learn a function that maps an input to an output. They have been widely applied to solve problems on ASV navigation, guidance and control.

\subsubsection{Multilayer Perceptron (MLP)}

MLP (also called FeedForward NN, that is, FFNN)
is one of the most basic neural network (NN) models that can approximate nonlinear functions and add one or multiple fully connected hidden layers between the output and input layers. MLP with more than one hidden layer is called DNN. Model training is usually performed through backpropagation for all layers.
MLP are usually used to estimate the model uncertainties of ASV controller. The modified version of MLP is listed as follows. 

\textbf{Radial Basis Function (RBF):} The RBF network
can effectively accelerate convergence speed and avoid local optima because it utilizes Gaussian functions $\varphi(\cdot)$ as activation functions in the hidden layer, which is a local approximation network. RBF has only one hidden layer, and there is usually one neuron in the output layer, which does not use the activation function.

\textbf{Wavelet Neural Network (WNN):} Combining the characteristics of wavelet transform and NN, WNN has advantages such as time-frequency localization, self-learning ability, faster convergence speed, and low false alarm rate. Using the structure of MLP or RBF, WNN
uses the wavelet function $f(.)$ as the active function.

\textbf{Fuzzy Neural Network (FNN):} Fuzzy logic 
captures the uncertainties associated with human reasoning based on ``degrees of truth” method, which humans are more easily understood compared to the conventional ``true or false” method. The integration of fuzzy logic and NN, i.e., using a NN to learn the parameters of a fuzzy logic system, creates an FNN,
which can model uncertainty and nonlinearly in complicated systems. 

\subsubsection{Convolutional Neural Network (CNN)}

The basic structure of CNN contains an input layer, several convolution layers, several pooling layers, a fully connected layer, and an output layer. In traditional DNNs, the neurons in different layers are fully connected. In contrast, the neurons in the convolution layers and pooling layers of CNNs are not fully connected to their forward layer. As a result, sparse interaction and parameter sharing are the two most important characteristics in a CNN learning process. Based on the basic idea of CNNs, advanced architectures are designed to solve three fundamental problems in the computer vision field, i.e., image classification, object detection, and image segmentation. 
Some popular architectures are briefly introduced below, which are used to perceive the surrounding environment through images, videos, or data collected in the maritime environment.

\textbf{CNN-based Classification:} Image classification aims to predict the label for the given image. Based on traditional CNNs, several improved architectures with more complex structures are proposed, including LeNet-5, 
AlexNet,
VGG, 
GoogLeNet, 
Inception Net,
ResNet,
and others. As the top competitors of the ILSVRC, these CNN architectures achieve high classification accuracy, and some of them even surpass human-level performance on the ImageNet database. 
These networks are widely used in many applications as the basic models for feature extraction or applied in object detection and segmentation tasks as the backbone network. 

\textbf{CNN-based Object Detection:}
Autonomous vehicles must identify and locate an object when perceiving their surroundings with sensors. In general, there are two types of object detection frameworks:

(1) The two-stage algorithm first extracts candidate regions that may contain the object and then determines whether the proposed regions contain the object. The first stage uses a method such as selective search or a region proposal network (RPN) to generate the regions of interest (ROIs). Then, another network, such as ResNet, is leveraged to classify the proposed regions. Popular two-stage object detection methods include R-CNN, 
Fast R-CNN, 
Faster R-CNN, 
and Mask R-CNN. 

(2) The one-stage algorithm directly classifies an object and predicts the object-bounding boxes for an image in one step. More specifically, an image is input, and the class probabilities and bounding box coordinates are learned simultaneously. The typical framework includes You Only Look Once (YOLO),
Single Shot Detector (SSD) 
and RetinaNet.

\textbf{CNN-based Segmentation:}
In some tasks, each pixel in captured images should be labeled with semantic tags to indicate which object the pixel belongs to. Image segmentation refers to the classification of every pixel in the image. 

Fully Convolutional Networks (FCN) 
are the first CNN-based network that can solve segmentation problems. FCN replaces the fully connected layers of CNN with convolutional layers and upsamples the high-level feature map into a heatmap, the size of which is the same as the original image. The heatmap shows the classification result for every pixel of the image. The pyramid attention network (PAN)
further improves FCN by proposing the feature pyramid attention (FPA) and global attention upsampling (GAU) modules to learn better feature representations by considering the global context information. Another commonly used segmentation method is U-Net,
which is structured like a capital letter \textbf{U}. The architecture of U-Net contains two paths. The first path is the contraction path, which is composed of a traditional stack of convolutional and max pooling layers to capture the context in the image. The second path is the symmetric expanding path, which is used to enable precise localization using transposed convolutions. Compared to FCN, U-Net fuses the features of the two paths on more levels. In terms of the feature fusion strategy, FCN uses additive fusion and U-Net uses channel-dimensional concatenation.

\subsubsection{Recurrent Neural Network (RNN)}
RNN can make use of historical information for the prediction of future behavior by passing all the information from the beginning to the current end of the model through the hidden layer.
RNNs perform very well when modeling time-dependent and sequential data tasks, but usually fail to converge during training because of problems related to vanishing and exploding gradients. 
As an improved version of RNN, the long-short-term memory (LSTM) has a specially designed memory network, including a cell, an input gate, an output gate, and a forget gate, 
to address the vanishing gradient problem that limits the use of traditional RNNs. Similar to LSTM, the gate recurrent unit (GRU) has two gates as opposed to three gates in an LSTM cell, which means that GRU has fewer training parameters than LSTM.

\subsection{Unsupervised Deep Learning Method}

Unsupervised deep learning method can learn patterns from untagged data.

\subsubsection{AutoEncoder (AE)}
AE 
is an unsupervised learning algorithm that is designed to reduce the dimensions of the data by learning the structure within the data and ignoring the signal noise. It has an input layer, a hidden layer that reconstructs the input data for compressed representation, and an output layer that reconstructs the input. The network should minimize the difference between the input and output vectors. Therefore, AE is applied to reduce the noise and dimensions of the original input in ASV navigation module.

\subsubsection{Generative Adversarial Networks (GAN)}
GAN
is an adversarial learning framework designed for data generation that has a generative model \textbf{G}, which generates samples similar to training data, and a discriminative model \textbf{D}, which estimates whether a sample comes from training data or \textbf{G}. The training process for \textbf{G} is to minimize the Jensen–Shannon (JS) divergence between the real data distribution and the generated data distribution. The training process for \textbf{D} is to distinguish between real data and generated data as much as possible. Furthermore, a deep convolutional GAN (DCGAN) is proposed to provide an experiment-based optimal hyperparameter set and training skills to improve the stability of a GAN training process. 
To further improve the stability of GANs from their structure and eliminate problems such as model collapse and vanishing gradient, Arjovsky \textit{et al.} \cite{MartinArjovsky2017WassersteinG} proposed the Wasserstein GAN (WGAN), which uses the Wasserstein distance instead of the JS divergence to measure the difference between the real and generated data distributions. GAN has been widely used to expand the dataset by generating fake images, which is very useful for the ASV navigation module if the training dataset is not enough.

\subsection{Reinforcement Learning Method}
Supervised and unsupervised learning is generally trained with labeled or unlabeled datasets that are provided in advance. Reinforcement Learning (RL)
is another type of machine learning technique that enables an agent to learn mapping from situations to actions by maximizing a scalar reward or reinforcement signal. The agent observes the state of the environment $s_t$ and takes an action $a_t$ at time $t$ in an environment. Then, the environment enters a new state $s_{t+1}$ and issues an immediate reward $r_{t+1}$ to the agent. Finally, the agent tries to learn to select actions that maximize the cumulative reward $Q(s_t,a_t)$, a.k.a. $Q$-value. 

\subsubsection{Deep Reinforcement Learning (DRL)}

For a complex environment, when the number of states and actions increases dramatically, the DL-based method can automatically extract features from very large states and action spaces with high dimensions, which introduces deep Q-networks (DQNs). The combination of RL and DL enables agents to construct and learn knowledge directly from raw sensors or image signals without any predefined features \cite{mnih2015human}. However, DQN is only applicable to discrete and low-dimensional action spaces. It cannot be straightforwardly applied to continuous domains, since it relies on finding the action that maximizes the action-value function, which in the continuous-valued case requires an iterative optimization process at every step. To address this problem, Lillicrap \textit{et al.} \cite{lillicrap2015continuous} presented the deep deterministic policy gradient (DDPG), which is a model-free, off-policy actor-critic algorithm that utilizes deep function approximators that can learn policies in high-dimensional and continuous action spaces.

\section{Deep Learning Driven Navigation System}
\label{sec:navigation}

%

\subsection{Definition and Key Problems}


An ASV navigation system provides environmental and self-state information to guidance and control systems based on collected data \cite{zheng2013survey,liu2016unmanned}. ASV usually has many sensors onboard that cumulatively collect very large amounts of data to monitor the performance of the ASV and the local environment, including the shore and static or moving obstacles \cite{chen2018development}. In addition, with the support of an advanced communication system, ASV can also receive information about the global environment, which covers large areas, such as an entire port with hundreds of ships. This type of information includes data from other ships collected by AIS \cite{tu2017exploiting}, synthetic aperture radar (SAR) images \cite{jin2020patch}, and images and videos taken by cameras on Autonomous Air Vehicle (AAV) \cite{gallego2018automatic} or other platforms \cite{shao2019saliency}. The key problems of ASV navigation are discussed below.

\subsubsection{Environmental Perception} Shipboard and remote sensors can provide information to ASVs for surrounding awareness to address external factors introduced by waves, currents, winds, and weather in a typical maritime environment and adapt to complex environments that include a large number of stationary or moving obstacles in real time \cite{elkins2010autonomous}. However, using highly dimensional data collected from a variety of types of equipment to reflect the complex environment is a very large challenge.

\subsubsection{State Estimation} An ASV state usually includes its position, orientation, velocity, and acceleration. 
However, the positions, Euler angles and linear and angular velocities of a ship measured by sensors such as global navigation satellite systems (GNSS) and inertial measurement units (IMUs) usually contain noise and errors, which can lead to failure. In recent years, advanced onboard sensors such as camera \cite{chen2018development,zhan2019autonomous,han2020autonomous}, radar, LiDAR \cite{yao2019lidar} and remote SAR image \cite{ball2017comprehensive} have improved the accuracy of the estimation. However, several unique situations, such as noise introduced by winds, waves, and weather, and harsh communication environments in the open sea, make accurate state estimation very challenging. 

\subsubsection{Data Processing Problem} The large number of data streams collected by heterogeneous sensors are different from each other in temporal and spatial resolution, data format, and geometric alignment. Therefore, certain processes, such as compression, dimensionality reduction, and fusion, may help improve the quality of data.



\subsection{Navigation with Deep Learning Model}

\subsubsection{Environmental Perception}
\label{sec_objectdetect}
Existing related work regarding the application of DL methods on maritime environmental perception can be categorized by the type of navigation data source, i.e., data collected by onboard or onshore equipment (optical and infrared images, radar images, and point clouds), or images captured by satellite (optical remote sensing images and SAR images). 
Here, based on different data source types, we introduce the DL-based environmental perception methods for optical and radar images, respectively.

\begin{figure}
\subfigure[Horizontal bounding boxes]{
\centering 
	\includegraphics[width=0.46\columnwidth]{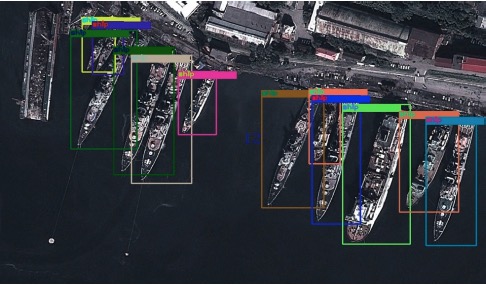}
}
\subfigure[Rotated bounding boxes]{
\centering 
	\includegraphics[width=0.46\columnwidth]{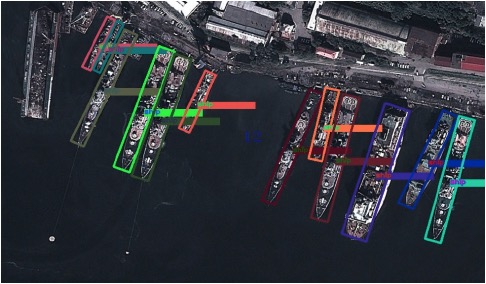}
}
\caption{Ship detection with optical remote sensing \cite{yang2018automatic}} 
\label{optical_ship} 
\end{figure}

\textbf{The Processing of Optical Images:} They are captured by on-board, onshore sensor, or satellite sensor, and usually experience the following challenges:


\begin{itemize}

\item The image quality depends greatly on the time of day and the absence of cloud coverage.

\item A large quantity of data has high resolution and is thus more difficult to utilize in real-time applications.

\item It is difficult to separate ships in ports because there are complex contexts such as buildings, docks, and overlapped ships that are usually closely docked side by side.


\item It is also difficult to locate a ship accurately because the target ships have extremely long and thin shapes and arbitrary rotations, as shown in \figurename~\ref{optical_ship}. 

\end{itemize}

Above challenges have been studied to detect ships with complex backgrounds \cite{nie2019deep,shao2019saliency,shan2020siamfpn,wu2020coarse,liu2021enhanced,lin2017fully,nie2020attention,yang2018automatic,yang2018position,huang2019ship}, detect multiscale ships \cite{shao2019saliency,shan2020siamfpn,wu2020coarse,yang2018automatic,yang2018position,li2020novel,yu2020cascade,zhang2020grs}, and detect ships with small data sets \cite{leclerc2018ship,moosbauer2019benchmark,chen2020deep}.  

For ship detection with complex backgrounds, extracting
appropriate feature representations that can better distinguish ships from surrounding turbulence is very important. Researchers have tried to integrate high- and low-level features for ship detection. One typical solution is to extract multiscale feature representations. Based on Mask R-CNN, Nie \textit{et al.} \cite{nie2020attention} added a bottom-up path to propagate low-level features to the top layer. Another solution is to extract features using image pyramids, the feature maps of which from low to high layers form a pyramid-like shape. Yang \textit{et al.} \cite{yang2018automatic,yang2018position} applied a dense FPN, in which the feature maps in different layers are densely connected and merged by concatenation. Huang \textit{et al.} \cite{huang2019ship} designed skip-connection path networks to extract features from each CNN layer and fuse all extracted features. Furthermore, expanding the training data set using a data augmentation strategy \cite{liu2021enhanced} and segmenting the sea and land areas before feature extraction \cite{shao2019saliency,wu2020coarse} are also effective strategies. 

For multiscale ship detection, popular object detection methods based on horizontal region detection have a large redundancy region for the bounding box of ship detection (see \figurename~\ref{optical_ship} (a)) compared to region detection based on rotated bounding boxes, which can be rotated according to the orientation of the ship (see \figurename~\ref{optical_ship} (b)). Li \textit{et al.} \cite{li2020novel} constructed two regression branches to independently predict the location of the center points $x, y$, the width $w$, and height $h$, and the orientation $\theta$ of the boundary box based on different CNN characteristics. For arbitrarily oriented ships, Yu \textit{et al.} \cite{yu2020cascade} proposed an anchor-assisted strategy that accurately predicts rotational bounding boxes and does not require manual anchor design. Zhang \textit{et al.} \cite{zhang2020grs} proposed an anchor-free rotation ship detection method by transforming the ship detection task into a binary semantic segmentation task, which directly detects pixels belonging to ships and predicts the distance between each pixel and four boundaries.

For ship detection with small dataset, to increase the training samples, researchers pre-trained a DL-based object detection framework on a well-established open dataset \cite{leclerc2018ship}, or generate fake ship images with GAN \cite{chen2020deep}.

\textbf{The Processing of Radar Images:} They are captured by on-board, onshore sensor, or satellite sensor, and usually suffer from the following issues:

\begin{figure}
\subfigure[Small and densely clustered ships with big bounding box]{
\centering 
	\includegraphics[width=0.46\columnwidth]{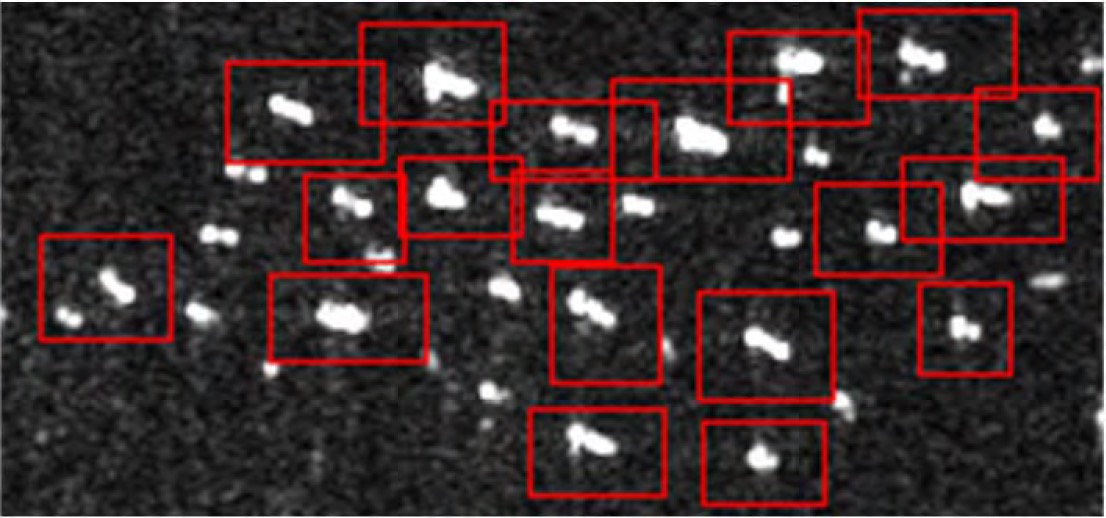}
}
\subfigure[False detection on the land]{
\centering 
	\includegraphics[width=0.46\columnwidth]{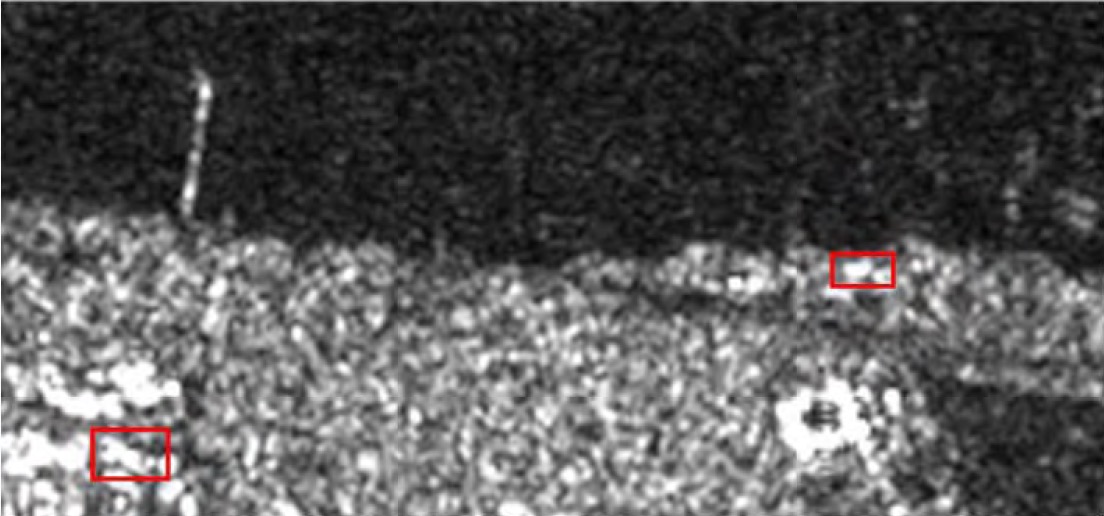}
}
\caption{Ship detection in SAR image \cite{zhao2018cascade}} 
\label{sar_ship} 
\end{figure}

\begin{itemize}

\item Although SAR is capable of working all day under any weather conditions, SAR images usually have a lower resolution and fewer pixels than optical remote sensing images. Thus, some of the models used on optical images cannot be applied directly to SAR images.

\item Small and densely clustered multi-scale ships in SAR images that occupy only a few pixels are very difficult to detect (see \figurename~\ref{sar_ship} (a)).

\item Ships in SAR images exist at a variety of scales, have arbitrary directions, and are densely arranged or even overlapped in ports.

\item Objects with analogical scatterings on land cause high false alarm rates (see \figurename~\ref{sar_ship} (b)).

\item Massive open SAR image datasets are rare and extensively labeled samples require a large amount of manual labor.

\end{itemize}

Above challenges have been studied to detect multiscale ship \cite{cui2019dense,kang2017contextual,zhao2020attention,zhao2018cascade,li2021novel,fu2020anchor,oksuz2020imbalance,johnson2019survey,zhang2021balance} and detect ship with small dataset \cite{ward2018ship,schwegmann2016very,li2017ship,lin2018squeeze,lin2018squeeze,deng2019learning,li2020sar}.

For multiscale ship detection, similar to optical images, integrating more extracted features from original SAR images to detect a ship is also useful. To obtain more semantic information, Cheng \textit{et al.} \cite{cheng2021robust} fused features extracted from radar and optical images to identify small objects on a water surface. VGG-13 and the backbone network of YOLOv4 were adopted to extract features from optical and radar images, respectively. Sometimes, it is very difficult to distinguish backscattering points of interference from actual ships in SAR images. Therefore, exploring the relationships between local features and their global dependences and re-inspection of information in different feature maps can improve the performance of multiscale ship detection in SAR images. Zhao \textit{et al.} \cite{zhao2020attention} proposed a two-stage detection method that adopts and combines a receptive field block and a convolutional block attention module to enhance the relationships of local features with their global dependence and boost significant information while suppressing interference. Furthermore, as a type of radar image, SAR images contain frequency information and can provide features in the frequency domain \cite{zhao2018cascade,li2021novel}. 

The lack of substantial ground truth data for training is one of the major problems in the detection and classification of ships and other objects in SAR images \cite{ward2018ship}, which has hindered the development of object detectors. In addition to obtaining more datasets, effective efforts have been made, including the addition of simulated SAR images to a training dataset, training deeper networks that extract higher levels of data presentation features \cite{schwegmann2016very}, and pre-training the object detection model on large open datasets \cite{li2017ship,lin2018squeeze}, have been made to reduce the false-positive rate on small datasets, improve detection accuracy, and overall performance. Furthermore, generating a fake dataset with WGAN can also improve the performance of ship detection in SAR images \cite{li2020sar}.

\subsubsection{State Estimation}
DNN-based methods can estimate the motion of ASVs in the presence of nonlinear dynamics and high-dimensional sensor data with different sampling frequencies. Typical tasks include ship behavior recognition \cite{chen2020video,chen2020ship}, position identification \cite{zhang2021multiscale} and motion prediction \cite{zhang2021multiscale,skulstad2020hybrid}.

\textbf{Ship Behavior Recognition:} High-fidelity ship kinematic information (displacement, movement speed, sailing angle, etc.) can be estimated directly from maritime surveillance video. Chen \textit{et al.} \cite{chen2020video} recognized ship behaviors based on video data in four steps: Ship feature extraction based on YOLO, bounding box generation, position identification based on geometry theory, and behavioral analysis. Chen \textit{et al.} \cite{chen2020ship} proposed a CNN ship movement mode classification algorithm to classify ship movements by converting AIS trajectories of a ship into images with different movements.

\textbf{Ship Position Identification:} DNN was trained based on historical position data estimated by a star sensor to predict and compensate for an INS navigation error when the star sensor is unusable \cite{wang2015performance}. 

\textbf{Ship motion prediction:} Zhang \textit{et al.} \cite{zhang2021multiscale} utilized LSTM to capture the inherent law of ship motion on each frequency scale. An attention mechanism was later adopted to further improve the accuracy of the prediction. 

\subsubsection{Data Processing}

It is very difficult to use the large amount of navigation data in maritime environments, which suffer from the noise introduced by limited on-board loads, poor communication, and unexpected situations. Therefore, DNN-based methods have been adapted to extract data representation before applying traditional multisensor methods to reduce data dimension or improve data quality \cite{naeem2012integrated}.

To reduce the data dimension, an autoencoder system architecture was proposed to reduce the performance dimensions and navigation data during transmission \cite{perera2017machine}. To improve the quality of the data, Cheng \textit{et al.} \cite{cheng2018concise} applied CNN in a data fusion module to extract joint information from different types of raw input, including the state of motion of the vessel, the state of existence of obstacles, and the previous control behavior. The output of a CNN model is a state vector StateID that contains the operating states of the vessel.

\section{Deep Learning Driven Guidance System}
\label{sec:guidance}

\subsection{Definition and Key Problems}
Global path planning and local path planning are two main problems for ASV guidance \cite{liu2016unmanned}. Traditionally, the path planning problem needs to be transformed into a solvable problem, such as a search problem, which can be solved by a deterministic method that guarantees the provision of a complete and consistent search result as long as it exists. Alternatively, this problem can be solved by a heuristic method that is able to obtain an approximate solution with a near optimal result if the deterministic approach is ineffective. These solutions are listed in Table \ref{path_planning_algorithm}.

\subsubsection{Global Path Planning}
Global path planning should generate an appropriate path from the original to the destination based on static obstacle map information or historical data. With an increasing amount of AIS data available, how to make the best use of these massive data to produce a global path for ASVs has become a challenge \cite{borkowski2017ship,wen2020automatic}. 

\subsubsection{Local Path Planning}
The predefined trajectory should be able to adjust according to tasks, complex and dynamic environments in real time, which requires local path planning. Currently, in congested sea areas such as complex ports or waterways, different surface vehicles frequently encounter each other. To avoid collisions in an efficient manner, ``own ship" (OS) and ``target ships" (TSs) should comply with widely accepted regulations such as COLREGS \cite{zhao2019colregs}. More specifically, if two ASVs encounter one another in a water way, from the first-person perspective, OS and TSs have four types of encounter situations, i.e., the head-on, stand on, give way and overtake encounters
. Before generating a local path, how to make appropriate decisions (i.e., choose the encounter situation) in real time is a very large challenge for researchers, especially under more complex situations such as OS encounters many TSs simultaneously.

\subsubsection{Constraints}
In real-world scenarios, certain constraints should be considered according to the specific task an ASV is assigned to \cite{zhou2020review}, such as: 

\begin{table}[]\caption{The category of path planning algorithm \cite{tam2009review}}
\resizebox{0.5\textwidth}{!}{
\label{path_planning_algorithm}
\begin{tabular}{lll}
\toprule
\textbf{General Methods} & \textbf{Specific Methods}   \\  \midrule
\textit{\textbf{Deterministic Searching Algorithm}}   & \\
Roadmap based searching    & Visibility graph \\
- Map construction methods & Voronoi diagrams \\
                           & Probability road map \\
Roadmap based searching    & A* searching \\
- Searching methods        & D* searching \\
                           & Field D* searching \\
Potential Field            & Conventional potential field \\
                           & Harmonic potential field \\
                           & Potential field by fast marching \\ 
Optimisation method        & Mixed integer programming \\
                           & Optimal control \\
\\
\textit{\textbf{Heuristic Searching Algorithm}}   & \\
Evolutionary algorithm    & Genetic Algorithm (GA) \\
                          & Particle swarm optimisation \\
                          & Asexual reproduction optimisation\\
                          & Ant Colony Algorithm (ACA)\\
Neural Network            & MultiLayer Perceptron (MLP) \\
                          & Fuzzy Neural Network (FNN) \\
                          & Long Short Term Memory (LSTM)\\
                          & Deep Reinforcement Learning (DRL)\\
\bottomrule
\end{tabular}
}
\end{table}

\begin{itemize}
\item \textit{Geography constraints} include seacoasts, rocks, and small islands that are present on geological maps. For global planning in a large area, such as navigating from one port to another, considering only geographical constraints is sufficient.

\item \textit{Shape constraints} refer to the specific size (length, width, and height) of the ASV, which must be considered when planning a path in the middle scale area, where an ASV cannot be treated as a point. For example, if ASV enters a channel, the width of the ASV and the channel will have a very large influence on the path planning process.

\item \textit{Kinematics constraints} represent the specific ranges of ship and acceleration velocities, which should be considered if the ASV enters an inner port.

\item \textit{ASV dynamic constraints} contain the inertia forces and moments of the surge, swaying, yawing, etc. of the ship, which cannot be neglected for precise path planning in small-scale areas, such as ASV berthing, which requires knowing how to precisely steer to berth and must consider kinematics and shape constraints.

\end{itemize}



\subsection{Guidance with Deep Learning Model}

\subsubsection{Global Path Planning}
Conventional deterministic and heuristic search algorithms can generate a collision-free path between the start point and the destination. However, in existing research, there are several complex tasks that the deterministic method cannot complete but can be solved by DNN. These include planning the most energy-efficient path \cite{zhang2019data}, predicting future paths based on historical trajectories \cite{tu2017exploiting,borkowski2017ship,wen2020automatic,gao2021novel}, and planning the optimal path for multiple tasks \cite{liu2018efficient}, which are each described below.  

\textbf{Planning the Most Energy Efficient Path:} It is very difficult to model the relationship between the environment and energy-efficient-related parameters in mathematical formulas. Zhang \textit{et al.} \cite{zhang2019data} proposed a data-driven ship speed optimization model and an ice route planning model to calculate the path of the ship with the highest energy efficiency for the Arctic area. On the basis of historical data, a DNN-based method was used to determine the relationship between ice concentration, ship speed, and an energy efficiency indicator, and then the optimal path and speed were calculated. 

\textbf{Predicting the Future Paths:} The future locations of ships can be predicted by DNN trained with turning point data such as latitude and longitude, speed, course, length, width, and draft of the ship \cite{wen2020automatic}. Instead of predicting the trajectory iteratively, Gao \textit{et al.} \cite{gao2021novel} performed multistep predictions by predicting both the support point and the destination point, where the destination point was generated by historical data, and the support point was produced by a trained LSTM model.

\textbf{Planning the Optimal Path for Multiple Tasks:} To visit multiple water monitoring stations on the surface with minimal costs, Liu \textit{et al.} \cite{liu2018efficient} mathematically summarized the task as the travel salesman problem, the goal of which is to visit all water monitoring stations and return to the starting point. In the global path planning stage, a self-organizing map-based DNN was applied to learn the relationship between the location of the water monitoring stations and the optimal execution sequence.

\subsubsection{Local Path Planning}
Conventional methods do not address sea or weather conditions or consider nonlinear ship dynamics due to the following shortcomings \cite{polvara2018obstacle}:

\begin{itemize}

\item The computational complexity of ship collision avoidance mathematical models is too high to be calculated in an assumed short period in highly dynamic environments with multiple objects.

\item Predefined architectures can only adopt some specified situations that involve oversimplified assumptions in risk assessment and ship dynamic constraints, which cannot adapt to complex encounter situations that need to comply with COLREGs.

\item The control law is usually formed as complicated formulas that consider all possible situations, which cannot be further adapted through changes.

\end{itemize}

Therefore, most of the previous studies address only one obstacle for one instance, which is not practical in increasingly busy maritime environments. In this section, we review the DRL-based \cite{shen2019automatic,zhao2019colregs,xie2020composite,chun2021deep} and DNN-based \cite{shen2019automatic,zhao2019colregs,gao2020ship} local path planning methods.

\textbf{DRL-based Local Path Planning:} A DRL model learns how to react through continuous interaction with an uncertain environment in real time. Ideally, a reward function should reward an agent for reaching a destination and avoiding collision with other objects while complying with COLREGs \cite{zhao2019colregs}. However, to avoid collisions, ASV must deviate from its path and sometimes even move in the opposite direction of the destination, which will cause penalties.  In \cite{shen2019automatic}, if the distances detected from other ships exceed the threshold value, a negative reward will be assigned; otherwise, a positive reward will be continuously provided. In \cite{zhao2019colregs,chun2021deep}, a path-following reward function is used in a safe navigation environment. If TSs enter a safe area around the OS, the collision avoidance reward function will be activated.

\textbf{DNN-based Local Path Planning:} In a practical environment, the OS encounters many TSs; thus, the number of states related to the TSs changes continuously. However, DNN has only a fixed-dimensional input. As a result, to handle the multi-ship encounter situation, in \cite{shen2019automatic}, the last five records of detected distances were used as input. In \cite{zhao2019colregs}, the input dimension of DNN is set to four by categorizing the TSs into four regions defined by COLREGs. 
Gao \textit{et al.} \cite{gao2020ship} combined LSTM and sequence conditional GAN to learn 12 types of ship encounter modes from AIS data and make an anthropomorphic decision to avoid ship collisions.

\section{Deep Learning Driven Control Systems} 
\label{sec:control}

\subsection{Definition and Key Problems}
\begin{figure}
\centering 
\includegraphics[width=0.43\textwidth]{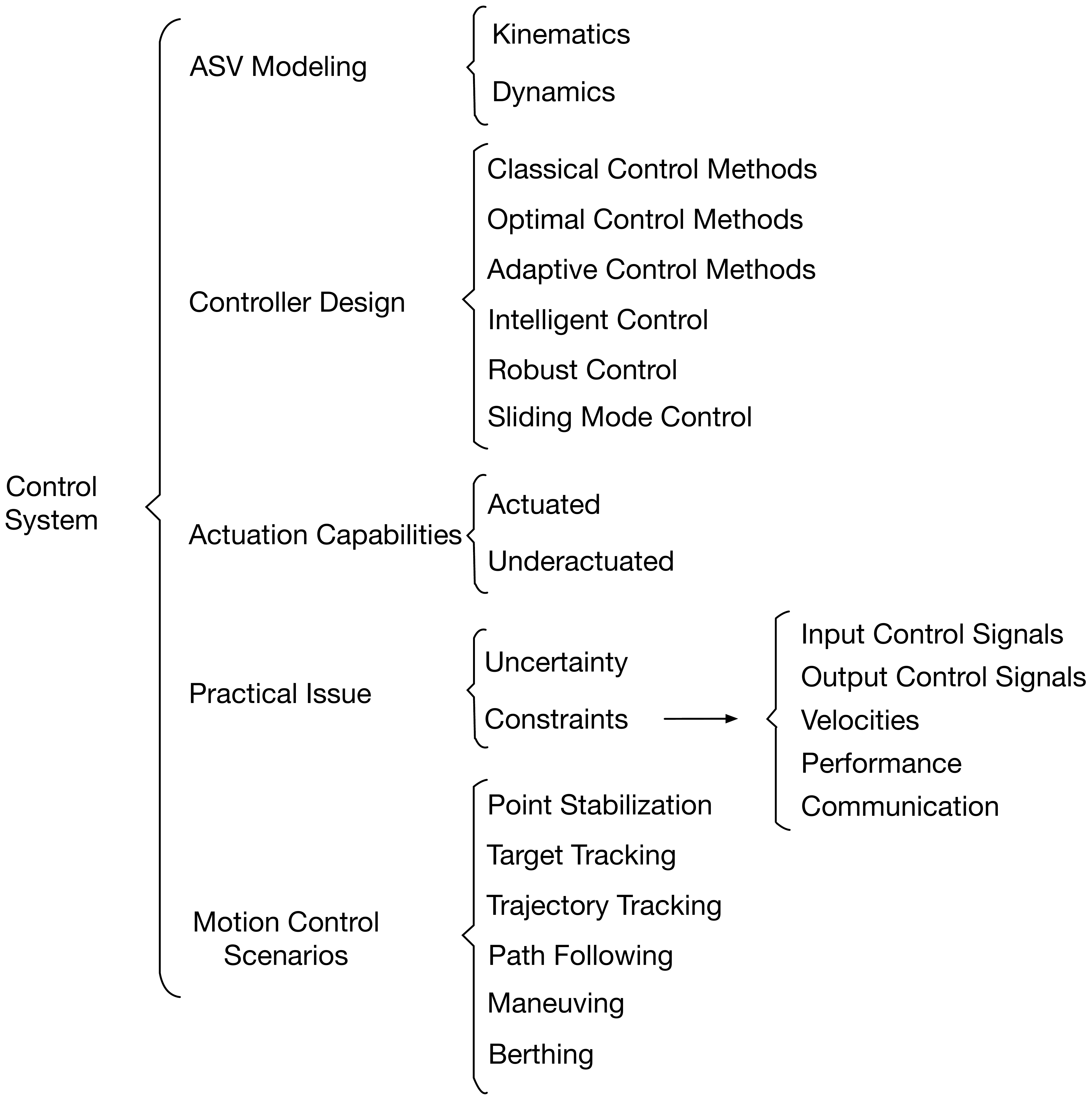} 
\caption{Problems considered in ASV controller design} 
\label{controller_issue} 
\end{figure}

A motion controller of a surface vehicle obtains input references from the guidance system, calculates the differences in the input references and the actual output values correspondingly, and gives commands to the activators such as thrusters and rudders. \figurename~\ref{controller_issue} shows the problems that existing studies generally focus on to design a proper ASV controller. The details are as follows.  


\subsubsection{ASV Modeling} 

ASV modeling requires the description of the motion of surface vehicles, which is divided into kinematics and dynamics. Kinematics only considers geometrical aspects of motion, and dynamics is the analysis of the forces causing motion. The vast majority of studies confine surface vehicles into three DOFs, since the roll and pitch motions are negligible in many water environments. As shown in \figurename~\ref{ship_DOF}, the moving coordinate frame that is attached to the vessel is called the body-fixed reference frame, $\sum x_by_bO_bz_b$.
The origin of the body-fixed frame is chosen to coincide with the center of gravity of the vessel. Moreover, the ASV kinematic model of the body-fixed frame is described relative to an inertial reference frame as follows:
\begin{eqnarray}\label{kinematicequation}
\dot{\bm{\eta}}=R(\eta)\bm{v},
\end{eqnarray}
where $\bm{\eta}=[x~y~\psi]^{T}\in\mathbb{R}^{3}$ denotes the position and heading angle of the vessel in the inertial frame, $\bm{v}=[u~v~r]^{T}\in\mathbb{R}^{3}$ denotes the surge velocity, the sway velocity and the angular velocity in the body-fixed frame, and $R(\eta)=[\cos{\psi}~  -\sin{\psi} ~  0;
\sin{\psi}~  \cos{\psi} ~ 0; 0~ 0~ 1]^{T}$ is the transformation matrix converting a state vector from the body-fixed frame to the inertial frame.

Following the notation developed by Fossen \cite{fossen1994guidance}, the nonlinear dynamics of surface vessels is described by the following differential equation.
\begin{eqnarray}\label{GeneralASVDynamics}
\bm{M}\dot{\bm{v}}+\bm{C}(\bm{v})\bm{v}+\bm{D}(\bm{v})\bm{v}=\bm{\tau}+\bm{\tau}_{\text{env}},
\end{eqnarray}
where $\bm{M}\in\mathbb{R}^{3\times3}$ is the positive-definite symmetric mass and inertia matrix, $\bm{C}(\bm{v})\in\mathbb{R}^{3\times3}$ is the skew-symmetric vessel matrix of Coriolis and centripetal terms, $\bm{D}(\bm{v})\in\mathbb{R}^{3\times3} $ is the positive-semidefinite drag matrix, $\bm{\tau}=[\tau_u~ \tau_v~ \tau_r]^{T}\in\mathbb{R}^{3}$  is the applied forces and torques generated by propellers in a body-fixed frame, and $\bm{\tau}_{\text{env}} \in\mathbb{R}^{3}$ is the environmental disturbances from the winds, currents and waves.

According to the thruster configuration of a vessel, the actuation force and moment vector $\bm{\tau}$ can be written as
\begin{eqnarray}\label{AppliedForceMaxtrix}
\bm{\tau}=\bm{B}\bm{u},
\end{eqnarray}
where $\bm{B}\in\mathbb{R}^{3\times {n_u}}$ is the control matrix that describes the thruster configuration and $\bm{u}\in\mathbb{R}^{{n_u}}$ is the control vector that represents the forces generated by the propellers, where $n_u$ is the dimension of the control vector.

Furthermore, by combining (\ref{kinematicequation}), (\ref{GeneralASVDynamics}) and (\ref{AppliedForceMaxtrix}), the complete dynamic model of the vessel is reformulated as follows: 
\begin{eqnarray}\label{MPCdynamics}
\dot{\mathbf{q}}(t)=f(\mathbf{q}(t),\mathbf{u}(t)),
\end{eqnarray}
where $\mathbf{q}=[x~y ~\psi~ u~ v~ r]^{T}\in \mathbb{R}^{6\times1}$ is the vessel state vector and $f(\cdot, \cdot, \cdot): \mathbb{R}^{n_q}\times \mathbb{R}^{n_u} \longrightarrow \mathbb{R}^{n_q}$ denotes the continuously differentiable state update function. The system model describes how the full state $\mathbf{q}$  changes in response to applied control input $\mathbf{u} \in\mathbb{R}^{{n_u}}$.


\subsubsection{Controller Design} 
A controller can be built based on classical, optimal, adaptive, intelligent, robust, and sliding mode control methods, or a combination of these techniques, as listed in Table \ref{control_algorithm}. 

\begin{table}[]\caption{Main control techniques \cite{azzeri2015review}}
\resizebox{0.5\textwidth}{!}{
\label{control_algorithm}
\begin{tabular}{lll}
\toprule
\textbf{Control Techniques} & \textbf{Examples}   \\  \midrule
Classical Control Methods  & Proportional-Integral-Derivative (PID) \\
Recursive Control & Backstepping \\
Adaptive Control &  NN Adaptive Control\\
Hierarchical Control System & \\
Intelligent Control        & Neural Networks (NN) \\
                           & Bayesian Probability\\
                           & Genetic Algorithms \\
                           & Fuzzy Logic \\ 
                           & Machine Learning \\
                           & Evolutionary Computation \\
Optimal Control            & Linear-Quadratic-Gaussian Control (LQG) \\
                           & Model Predictive Control (MPC) \\
Robust Control             & H-infinity Loop-Shaping \\
                           & Sliding Mode Control (SMC) \\
                           & Dynamic Surface Control (DSC)\\
Stochastic Control         &  \\
Energy-shaping Control     &  \\
Self-organized Criticality Control    &  \\
\bottomrule
\end{tabular}
}
\end{table}

\subsubsection{Actuation Capabilities} A vehicle with full actuation can control all its DOFs simultaneously independently; otherwise, the vehicle is underactuated. As the most common configuration among ASVs, underactuation makes the design of controllers much more difficult compared to actuated ASVs because only the surge and yaw axes are directly actuated with propellers and rudders. There are no actuators for direct control of sway motion for actuated ASVs. 

\subsubsection{Motion Control Scenarios}
\begin{figure}
\centering 
\includegraphics[width=0.45\textwidth]{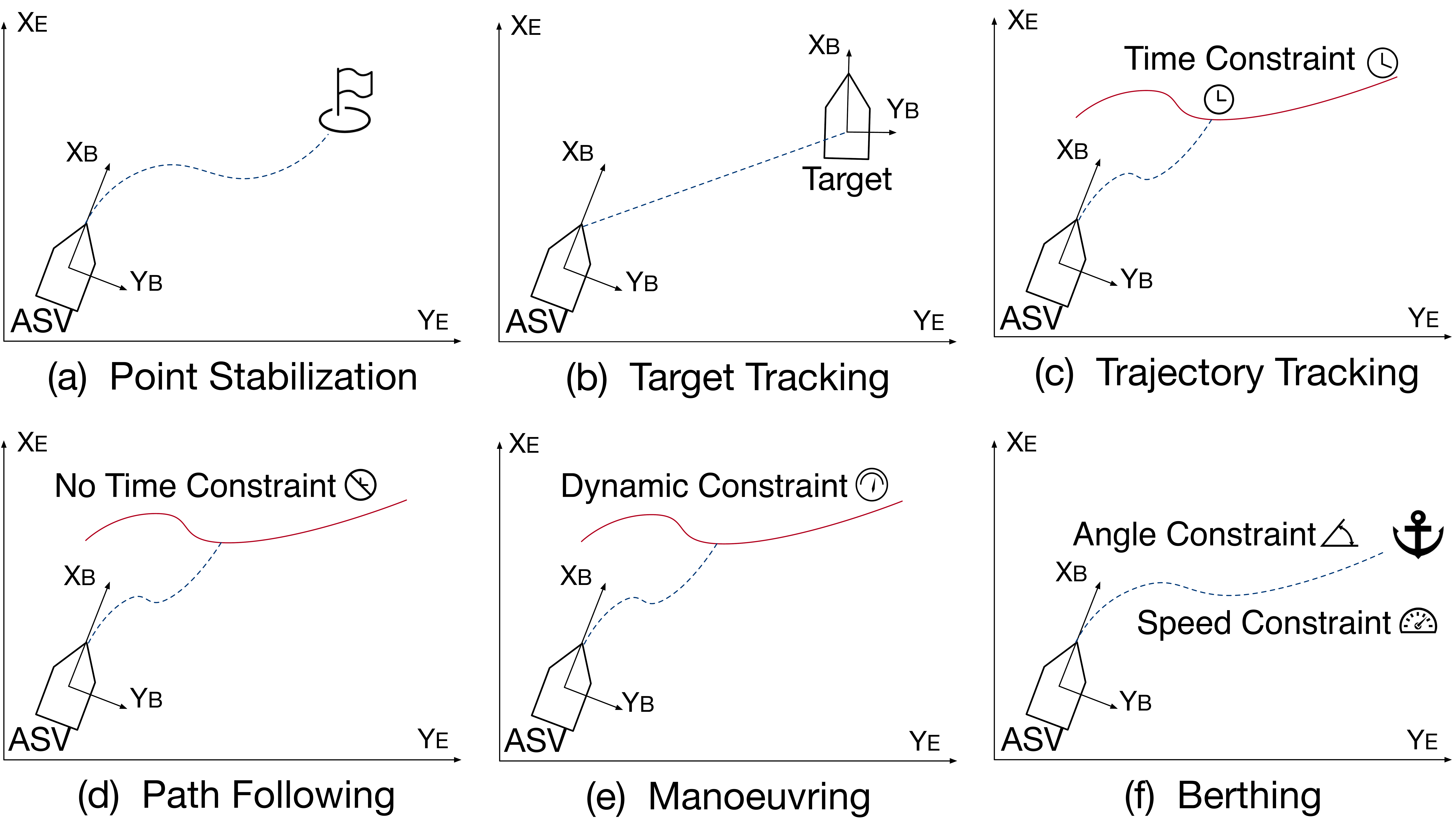} 
\caption{Motion Control Scenarios, (a)-(d) refer to \cite{peng2020overview}} 
\label{motion_control_tasks} 
\end{figure}

Many motion controllers are designed to solve one or more specific problems in different scenarios to accomplish their specified tasks. In general, depending on what motion information and constraints are available as \textit{a priori}, they can be categorized into point stabilizing, target tracking, trajectory tracking, path following, maneuvering, and berthing. The schematic diagrams are shown in \figurename~\ref{motion_control_tasks}, and we describe each motion control scenario below.

$\bullet$ Point Stabilization: The objective is to position and orient the ASV in fixed target operations without time constraints under changing ocean disturbances \cite{sorensen2011survey}. For underactuated ASVs, only discontinuous or smooth time-varying control is possible if all three coordinates are stabilized \cite{ashrafiuon2010review}. The thruster-assisted position mooring system (PM) for anchored vessels and the dynamic positioning system (DP) for free-floating vessels are two main types of positioning systems for traditional vessels \cite{sorensen2011survey}. PM consumes less energy because it usually has many anchor lines for each platform. The DP system uses only actuated thrusters to accurately maintain the position and heading of the ASV at a fixed location or on a predetermined track. DP has gained increasing attention because it easily changes position and is adapted to various ocean environments. 

$\bullet$ Target Tracking: The goal is to track the motion of static or dynamic targets without knowing future information about the motion of the target \cite{breivik2008formation}. Related spatio-temporal constraints must be considered simultaneously.

$\bullet$ Trajectory Tracking: This refers to the ability to follow a specified path at a desired forward speed with a time constraint. Therefore, it is possible to consider the spatiotemporal constraints related to the target separately \cite{breivik2008straight}.

$\bullet$ Path Following (also called track keeping, path keeping, or course keeping): The objective is to follow a predefined path with constant forward velocity that only involves a spatial constraint \cite{breivik2008straight}. 

$\bullet$ Manoeuvring: As a subset of the path following \cite{peng2020overview}, the objective of maneuvering is to steer the ASV along a predefined path, while controlling speed can be addressed as a separate task \cite{skjetne2005adaptive}. The maneuvering problem can be divided into geometric and dynamic tasks. The first is similar to the following path, and the second assigns a time, speed, or acceleration along the path. In contrast, the tracking problem merges the above two tasks into a single task.

$\bullet$ Berthing: A ship should be able to stop near the berthing point at low speed \cite{zhang1997multivariable}.
As one of the most difficult problems in automated ship control, it is almost impossible to consider all possible situations during berthing due to the influence of unpredictable environmental disturbances. In more extreme cases, with drastic reduction in ship maneuverability, the signals of ship motion and noise are almost the same, making it difficult to adjust the rudder angle.

\subsubsection{Practical Issue} Model uncertainties and constraints are commonly considered issues. To address nonlinear systems with uncertainties and constraints, model-free methods such as DNNs and fuzzy logic controllers are widely used.

\textbf{Model Uncertainties:} They are usually caused by unknown dynamics \cite{qiu2019adaptive,shen2020mlp,zhang2020neuro}, underactuated ASVs \cite{dai2018adaptive}, high-speed maneuvering situation \cite{zhang2011nnffc}, sensor errors \cite{shen2020mlp}, or environmental disturbances \cite{zhang2021model}, which may introduce unknown parameters, terms, or functions into an ASV control system. 


\textbf{Constraints:} Various constraints widely exist in most physical systems, such as input / output control signals, velocities, performance and communication constraints \cite{zhao2013adaptive}. Control systems must address these limits and constraints, or the control performance will degrade, leading to control failure or even potential collisions.


$\bullet$ Input Control Signal Constraints: In practical control systems, feedback control systems are inevitably subject to actuator saturation/input saturation, which means that the control torques are constrained due to the physical limitations of actuators. If the control signals generated by the controller exceed a certain range, the tracking performance for the closed-loop system cannot be guaranteed.

$\bullet$ Output Control Signal Constraints: ASV outputs are not allowed to exceed a certain constrained distance from the predefined path; that is, the range of the output error is limited \cite{dai2018adaptive}. Such constraints are crucial for system performance and ASV safety, especially in narrow waterways.

$\bullet$ Velocities Constraints: In certain real scenarios, an underactuated ship moves in the open sea with limited forward and angular velocity. Therefore, velocity constraints are considered in some studies \cite{pan2015biologically}.

$\bullet$ Performance Constraints: To achieve steady-state tracking performance, the prescribed transient performance (e.g., predetermined convergence rate) is important in many practical applications \cite{he2018asymptotic}. For example, angle and LOS range errors should always stay within predefined regions.

$\bullet$ Communication Constraints: Communication resources are limited in a real maritime environment. Therefore, event-triggered control methods, which only update actuator state if a particular event occurs or a given condition is met, have been studied by many researchers \cite{zhang2020robust,zhang2020composite,li2020adaptive}. 

\subsection{Control with Deep Learning Model}

DNN and DRL are model-free estimators that map conditions to actions and are widely employed to develop robust adaptive controllers for uncertain nonlinear systems to estimate model uncertainties or generate a control signal. All controllers based on the DL model or related to the DL model are compared in Table \ref{nn_controllers} and categorized into 6 typical motion control scenarios. Several key works are described in the following paragraph for different purposes of the DL model.

\begin{table*}[]
\caption{The Comparison of DL-Based or Related Controllers}
\resizebox{\textwidth}{!}{
\label{nn_controllers}
\begin{tabular}{|l|l|l|l|l|l|l|l|}
\hline
\textbf{Applied DL} & \textbf{DL App.} & \textbf{Main Control Techniques} &\textbf{Proposed Model} & \textbf{DIS.} & \textbf{CON.} & \textbf{ACT.} & \textbf{REF.}                      \\ \hline
\multicolumn{8}{|l|}{\textit{\textbf{Point Stabilization}}}                                                                                                               \\ \hline
~~MLP & E.  &Backstepping& Robust adaptive position mooring control &\checkmark&In.&-&\cite{chen2012robust} \\ \hline 
~~RBF & E. &Backstepping& Robust adaptive nonlinear controller &\checkmark&-&-& \cite{du2013robust}\\ \hline
~~RBF & E.   &Backstepping& Robust adaptive output feedback control scheme &\checkmark&-&-& \cite{du2015adaptive}\\ \hline
~~RBF & E.  &Backstepping& Robust neural event-triggered control &\checkmark&Com.&Act.&\cite{zhang2020robust}\\ \hline

\multicolumn{8}{|l|}{\textit{\textbf{Target Tracking}}}                                                                                                                                                                                                                                                                                                       \\ \hline
~~MLP & E.   &ESO& Target tracking controller &\checkmark&In.&Un.& \cite{liu2018bounded}  \\ \hline
\multicolumn{8}{|l|}{\textit{\textbf{Trajectory Tracking}}}                                                                                                                                                                                                                                                                                                       \\ \hline

~~MLP & E. &Backstepping& Stable tracking controller &\checkmark&-&Act.&\cite{tee2006control}\\ \hline
~~RBF & E.  &Backstepping& Robust adaptive tracking controller &\checkmark&In.&-& \cite{chen2009neural}\\ \hline
~~WNN & G.   &WNN& Neural and auxiliary compensation controller &\checkmark&-&-& \cite{chen2011intelligent} \\ \hline
~~MLP & E. &NN\&PD& Adaptive output feedback controller &\checkmark&-&-& \cite{zhang2011nnffc}\\ \hline
~~RBF & E.  &Backstepping& Adaptive NN controller &\checkmark&Out.&Act.& \cite{zhao2013adaptive}\\ \hline 
~~One-layer & E.  &Backstepping& NN controller &\checkmark&-&-& \cite{pan2013efficient}\\ \hline     
~~RBF & E.  &Backstepping& Adaptive output feedback NN tracking controller &\checkmark&-&Act.& \cite{dai2014learning}\\ \hline
~~One-layer & E. &Backstepping& NN based tracking controller &\checkmark&In.Vel.&Un.&\cite{pan2015biologically}\\ \hline
~~FNN & E.  &SMC& Adaptive robust fuzzy neural controller &\checkmark&-&-&\cite{wang2014self}\\ \hline
~~MLP & E. &NN& Saturated neural adaptive robust controller &\checkmark&-&Un.&\cite{shojaei2015neural}\\ \hline
~~RBF & E. &HGO& Adaptive NN controller &&Out.&Act.&\cite{he2016adaptive}\\ \hline 
~~RBF & E. &NN&Adaptive output feedback controller &\checkmark&-& Un.& \cite{park2017neural} \\ \hline
~~RBF & E. &Backstepping& Trajectory tracking controller &\checkmark&-& Act. & \cite{zheng2017trajectory} \\ \hline
~~RBF & E.  &Backstepping& Sign of the error based adaptive NN controller &\checkmark&Per.&Act.& \cite{he2018asymptotic} \\ \hline
~~RBF & E.  &DSC \& backstepping& Adaptive neural controller &\checkmark&Per.&Un.& \cite{dai2018adaptive} \\ \hline
~~RBF & E.  &SMC \& backstepping &Adaptive sliding mode controller &\checkmark&In.&Un.& \cite{qiu2019adaptive} \\ \hline
~~RBF & E.  &SMC \& DSC& Adaptive dynamic surface controller &\checkmark&In.& Act. & \cite{shen2020mlp} \\ \hline
~~RBF & E.  &Backstepping \& HGO& Neuro-adaptive trajectory tracking controller &\checkmark&-& Un. &\cite{zhang2020neuro}\\ \hline
~~DRL & G.  &DRL& Mode-reference RL controller &\checkmark &-&-& \cite{zhang2021model}\\ \hline
~~RBF & E.  &DSC& Robust adaptive controller &\checkmark&In.Out.&-& \cite{ZHU2020robust}\\ \hline
~~RBF & E.  &DSC& Adaptive NN controller &\checkmark&In.&Un.& \cite{rout2020modified}\\ \hline
~~DRL & G.  &DRL& Actor-critic NNs based controller &\checkmark&Out.&Un.& \cite{zheng2020reinforcement}\\ \hline
~~RBF & E.  &Backstepping  & Robust adaptive controller &\checkmark&-&Un.& \cite{zhang2020robusttra}\\ \hline
~~RBF & E.  &Backstepping  & Adaptive neural output feedback controller &\checkmark&-&-& \cite{zhu2021adaptive}\\ \hline
~~RBF & E.  &SMC & BF-based adaptive NN SMC &\checkmark&-&-& \cite{yan2021barrier}\\ \hline
~~DRL & G.  &DRL& Data-driven performance-prescribed RL controller &\checkmark&Per.&-& \cite{wang2021data}\\ \hline
~~RBF & E.  &SMC& Fixed-time SMC &\checkmark&Vel.&Un.&\cite{zhou2021fixed}\\ \hline
~~MLP & E.  &MPC& PWM-driven model predictive controller &\checkmark&-&-& \cite{peng2021pwm}\\ \hline
~~MLP\%DRL & E.\&G. &DRL& RL based optimal tracking controller&\checkmark&Vel.&-&\cite{wang2021reinforcement}\\ \hline

\multicolumn{8}{|l|}{\textit{\textbf{Path Following}}}                                                                                            \\ \hline

~~RBF & G. &RBF& Ship steering control system& &-&-&\cite{unar1999automatic}\\ \hline
~~RBF & E.  &DSC& Adaptive NN path-following controller&\checkmark&-&Un.& \cite{zhang2013concise}\\ \hline
~~RBF & E.  &Backstepping& Robust adaptive RBFNN controller &\checkmark&In.&-&\cite{zheng2016path}\\ \hline 
~~RBF & E.  &DSC& Adaptive NN-DSC controller &\checkmark &In.&Un.&\cite{liu2017adaptive}\\ \hline
~~RBF & E. &RBF& Robust neural path-following controller &\checkmark&Com.&Un.&\cite{zhang2017robust}\\ \hline
~~DRL & G. &DRL& DRL-based path following controller&\checkmark&-&-&\cite{woo2019deep}\\ \hline
~~RBF & E.  &Backstepping& Adaptive NN event-triggered controller &-&-& Un.& \cite{li2020adaptive} \\ \hline
~~DRL & G.  &DRL& RL-based controller & &-&Un.& \cite{meyer2020taming} \\ \hline
~~DRL & G.  &DRL& Smoothly-convergent DRL method &\checkmark&-&Un.& \cite{zhao2020path} \\ \hline
~~RBF & E.  &DSC& Composite neural learning fault-tolerant controller&\checkmark&Com.&Un.&\cite{zhang2020composite}\\ \hline
~~Projection NN & G. &MPC& Quasi-infinite horizon MPC controller &\checkmark&Vel.&Un.&\cite{liu2020model}\\ \hline
~~RBF & E. &DSC\&backstepping & Adaptive NN control &\checkmark&Out.&Un.&\cite{rout2020sideslip}\\ \hline
~~Critic NN & E. &Backstepping & Dynamic path-following controller &\checkmark&Vel.&Un.&\cite{zhou2021event}\\ \hline

\multicolumn{8}{|l|}{\textit{\textbf{Manoeuvring}}}                                                                                                                                                                                                                                                                                                       \\ \hline

~~RNN & G.  &RNN& RNN manoeuvring simulation model & &-&-& \cite{moreira2003dynamic} \\ \hline
~~MLP & E.  &PID/PD& Self-tuning NN based PID controller & &-&-& \begin{tabular}[c]{@{}l@{}}\cite{fang2010application}\\ \cite{fang2012applying} \end{tabular} \\ \hline
~~RBF & G.  &RBF& Course keeping and roll damping controller &\checkmark&-&-& \cite{wang2017unscented} \\ \hline
~~LSTM & E. &LSTM& DL based dynamic model identification method & &In.&-&  \cite{woo2018dynamic}\\ \hline
~~DRL & G. & DRL & Concise DRL obstacles avoidance control & \checkmark & Vel. & Un. & \cite{cheng2018concise} \\ \hline

\multicolumn{8}{|l|}{\textit{\textbf{Berthing}}}                                                                                                                                                                                                                                                                                                       \\ \hline

~~MLP & G. &MLP& Multi-variable Neural Controller &\checkmark&-&-&\cite{zhang1997multivariable}\\ \hline
~~MLP & G. &MLP\&PD& NN and PD controller &\checkmark&In.&-& \cite{ahmed2013automatic} \\ \hline
~~MLP & G. &MLP& NN controller & &-&-& \cite{im2018artificial} \\ \hline
~~RBF & E.  &DSC& Auto-berthing control scheme &\checkmark&In.&Un.& \cite{qiang2019adaptive} \\ \hline
~~MLP & G. &MLP& NN based automatic ship docking &\checkmark&Vel.&-& \cite{shuai2019efficient} \\ \hline
\end{tabular}
}

\footnotesize{Notes: 1. (-) not mentioned; (\checkmark) considered. 2. (DL App.) Applications of DL, including Model Uncertainties Estimation [E.], Control Signals Generation [G.]; (DIS.) Disturbance; (CON.) Constraints, including constraints on Input control signal [In.], Output control signal [Out.], Velocities [Vel.], Performance [Per.], Communications [Com.]; (ACT.) Actuated [Act.] or Underactuated [Un.]; (REF.) References.}
\end{table*}

\subsubsection{Model Uncertainties Estimation} The unknown parameters, terms or functions introduced by unknown dynamics, underactuated ASVs, high-speed maneuvering situations, sensor errors, or environmental disturbances can be estimated by DNN.  

$\bullet$ Point Stabilization: DNN has been applied to estimate the model uncertainties for both PM systems \cite{chen2012robust} and DP systems \cite{du2013robust,du2015adaptive,zhang2020robust}. Actuator faults are frequently encountered for DP ships with multiple actuators. Zhang \textit{et al.} \cite{zhang2020robust} designed a DP control method that considers actuator faults and limited communication resources. The uncertainty of these subsystems was approximated by RBF, and the computational complexity was reduced by DSC, which greatly reduced gain-related adaptive parameters.




$\bullet$ Target Tracking: Without knowing the target velocity information, Liu \textit{et al.} \cite{liu2018bounded} used the extended state observer (ESO) and DNN to estimate the dynamics of the target and follower, respectively. Furthermore, the control torques are bounded by integrating the neural estimation model and a saturated function.

$\bullet$ Trajectory Tracking: DL model is capable of learning the unknown dynamics for fully actuated ASVs \cite{tee2006control,chen2009neural,zhao2013adaptive,pan2013efficient,wang2014self,he2016adaptive,he2018asymptotic,qiu2019adaptive,shen2020mlp,ZHU2020robust,zhu2021adaptive,zhang2021model,peng2021pwm,yan2021barrier,zhang2011nnffc,wang2021reinforcement}, estimating the unknown dynamics for underactuated ASVs \cite{pan2015biologically,shojaei2015neural,park2017neural,dai2018adaptive,zhang2020neuro,rout2020modified,zheng2020reinforcement,zhang2020robusttra,zhou2021fixed}
and approximating unknown disturbances \cite{tee2006control,chen2009neural,zhao2013adaptive,wang2014self,zheng2017trajectory,zheng2020reinforcement,yan2021barrier}.








DNNs are usually adopted to estimate the term of the control law, which is formed by unknown parameters, such as the inertia matrix $M$, the Coriolis and centripetal terms matrix $C(v)$, the damping matrix $D(v)$, and the unknown vector of gravitational and buoyancy forces and moments $g(\eta)$ \cite{tee2006control,chen2009neural,zhao2013adaptive,pan2013efficient,dai2014learning,wang2014self,he2016adaptive,qiu2019adaptive,shen2020mlp,ZHU2020robust}. DNNs can also estimate unmeasurable velocities \cite{zhu2021adaptive}. 

$\bullet$ Path Following: To address arbitrary uncertainties, DNN is used to approximate unknown dynamics \cite{zhang2013concise,zheng2016path,liu2017adaptive,zhang2017robust,zhang2020composite,zhang2017robust,li2020adaptive,rout2020sideslip,zhou2021event} and disturbances \cite{zheng2016path,liu2017adaptive}. For most of the existing work, researchers focused on controlling underactuated ASVs because they are characterized by strong nonlinearity, uncertainty of the model parameter, and constraints of the control input saturation, and are easily influenced by external interference \cite{zhang2013concise,li2020adaptive,liu2017adaptive,zhang2017robust,li2020adaptive,zhang2020composite,rout2020sideslip,zhou2021event}. 

In DNN-based backstepping design, Li \textit{et al.} \cite{li2020adaptive} combined a fast power reaching law with RBFNN in the controller design to accelerate the convergence rate of tracking error and achieve finite-time stabilization of the controller. The DSC technique can reduce the computational burden introduced by backstepping. In the DNN-based DSC design, DNN estimated the unknown system functions of the proposed controller to develop uncertain nonlinear multi-input-multiple-output (MIMO) time-delay systems \cite{zhang2013concise,liu2017adaptive}. 

$\bullet$ Manoeuvring: Among the 6-DOF of a ship, the roll motion has gained the most attention because large rolling motions may capsize a ship. Therefore, DNN was applied to estimate the unknown dynamics to reduce the rolling motion \cite{fang2010application,fang2012applying}. 

$\bullet$ Berthing: When approaching a port, a ship is very difficult to control and is easily influenced by disturbances and winds at a very low speed, making prediction or representation using differential equations difficult because the signal-to-noise ratio is too low for any controller to separate it from the real motion of the ship \cite{zhang1997multivariable}. The automatic berthing controller is a complicated MIMO system that can simultaneously evaluate many factors, such as the current speed of the ship, the angle of berth and the distance to the pier. DNN can be used to reconstruct uncertain modal dynamics and unknown disturbances \cite{qiang2019adaptive} or learn any nonlinear MIMO system and respond to any unknown situation if enough input and output data are available for training \cite{zhang1997multivariable,ahmed2013automatic,im2018artificial,shuai2019efficient}.

\subsubsection{Control Signals Generation} Data driven models such as DNN and DRL can generate control signals without an \textit{a prioir} model \cite{peng2020overview}.

$\bullet$ Trajectory Tracking: Wang \textit{et al.} \cite{wang2021data} proposed an RL control algorithm with a DNN-based actor-critic structure to establish a data-driven method-based optimal control scheme, which only requires ASV input-output data pairs. Here, the critic and actor DNNs learned the optimal policy and cost function simultaneously.

$\bullet$ Path Following: Without dependency on prior knowledge of dynamic modeling, DNN \cite{unar1999automatic} 
and DRL \cite{woo2019deep,zhao2020path,meyer2020taming} can generate the control signal directly. DRL can learn from the interactions between the agent and the environment to find the best policy without knowing any information in advance. Based on a two-DQN structure, Zhao \textit{et al.} \cite{zhao2020path} reduced the complexity of the control law by designing computationally efficient exploring and reward functions. 

$\bullet$ Manoeuvring: To adapt to various and complex navigation requirements, Cheng \textit{et al.} \cite{cheng2018concise} used the DRL method with a DQN architecture to control underactuated ASVs. The objectives and constraints, including destination, obstacle avoidance, target approach, speed modification, and attitude correction, were considered in the avoidance reward function, the input of which is provided by a CNN-based data fusion module, and the outputs of the DRL network were the propulsion surge force and the yaw moment.

$\bullet$ Berthing: Im \textit{et al.} \cite{im2018artificial} proposed an artificial NN controller that could automatically control the ship during berthing at the original port and other ports. The initial conditions of the inputs in the head-up coordinate system of other ports should be similar to those of the training data in the original port. Then, a DNN controller can adapt to different ports without retaining based on two key inputs, that is, the relative bearing and distance from the ship to the berth.

\section{Deep Learning in Cooperative Operations}
\label{sec:DL_fleet}

To carry out complex and large-scale missions in maritime cooperation scenarios, cooperative control of multiple ASVs offers increased efficacy, performance, scalability, and robustness and the emergence of new capabilities \cite{peng2020overview}. 
Furthermore, as an indispensable part of future transportation systems, autonomous systems are supported by the internal and external Internet of Things (IoT), big data platforms, and communication infrastructures. As shown in \figurename~\ref{fleet}, taking advantage of advanced communication technologies, together with multiple underwater and air vehicles, intelligent cooperative maritime operations have emerged. A considerable portion of ship intelligence will consist of a DL-based framework, especially for networks that are trained by image-based information and navigational actions rather than system parameters \cite{perera2020deep}. In this section, several issues regarding cooperative operations based on the application of DL methods are discussed. Table \ref{nn_coop_controllers} compares coordinated control methods based on DL or related to DL.

\label{fleet}
\begin{figure}
\subfigure[Typical scenarios \cite{breivik2008formation}]{
\centering 
	\includegraphics[width=0.46\columnwidth]{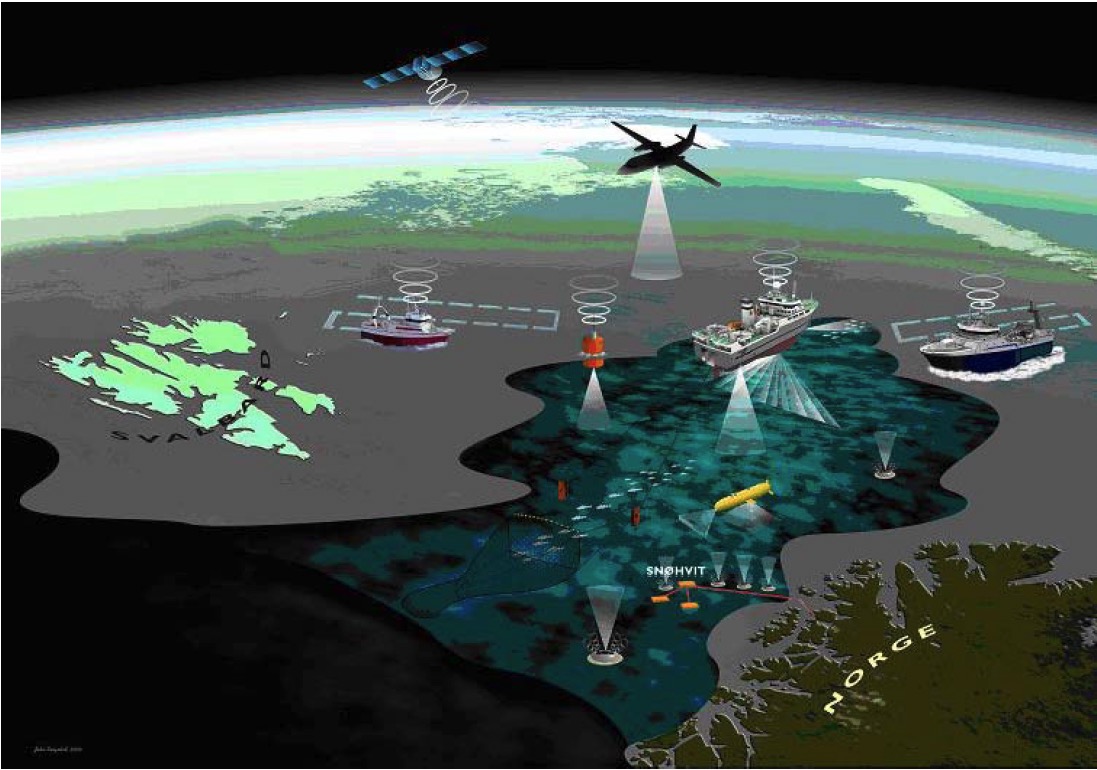}
}
\subfigure[Key techniques \cite{zolich2019survey}]{
\centering 
	\includegraphics[width=0.46\columnwidth]{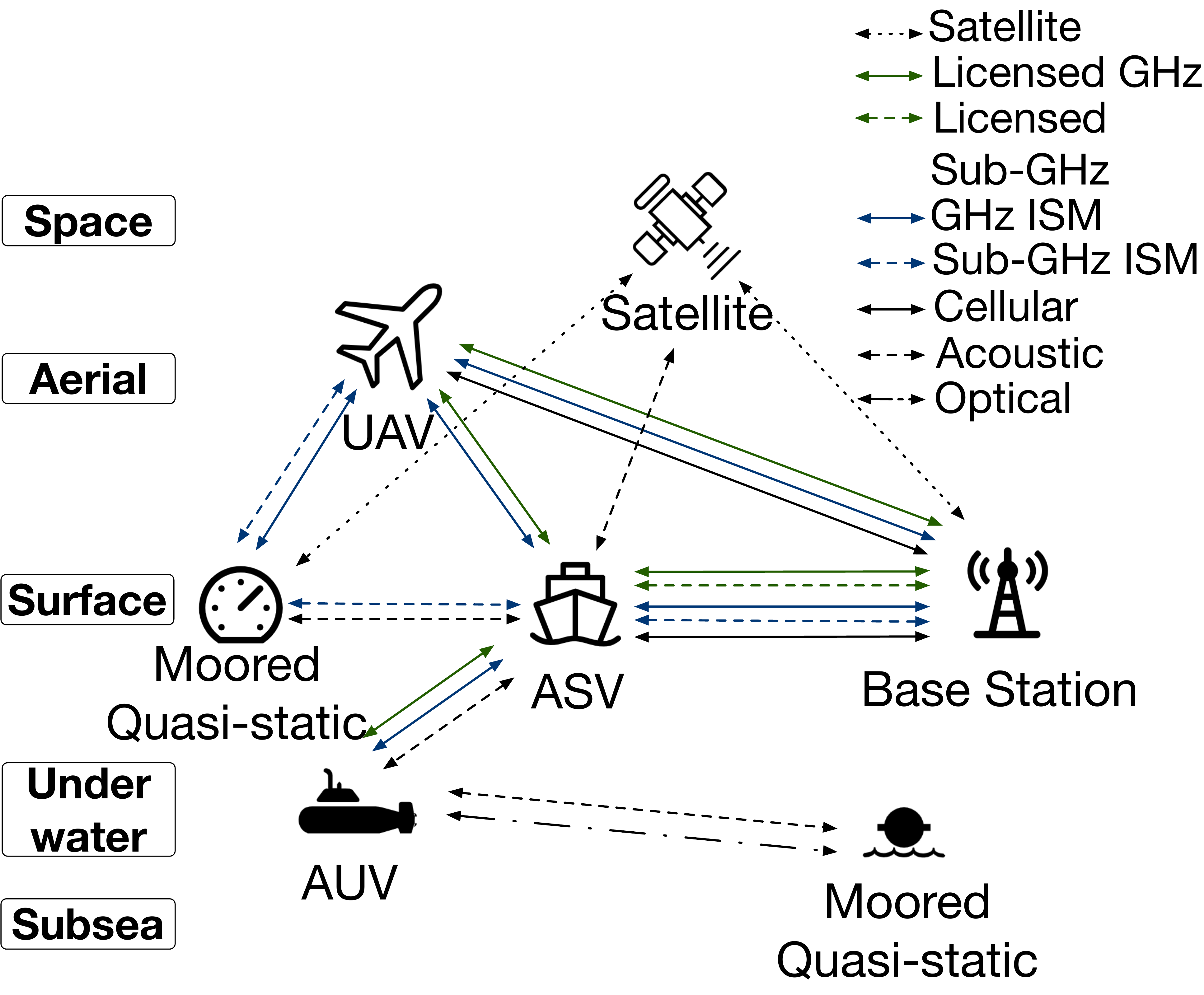}
}
\caption{Communication and Networking of Autonomous Systems } 
\label{fleet} 
\end{figure}

\subsection{Cooperative Control}

Cooperative control aims to force a group of ASVs to achieve and maintain the desired formation geometry by designing path-following, trajectory-following, and target-following controllers while ensuring that the agents complete the predefined task. Leader-follower formation control and leaderless formation control are two common strategies of cooperative control methods. The difference between these methods lies in whether the group follows a physical leader ASV or reaches a common value through local interaction \cite{peng2013leaderless}. In this survey, existing results can also be classified into coordinated controls guided by a trajectory, a path and a maneuvering guide according to various types of reference signals, which can be further classified into three architectures according to the available communication bandwidths and sensing abilities, i.e., centralized, decentralized and distributed controls. In addition to sophisticated guidance and control methods, the challenges of cooperative control systems with multiple ASVs include model uncertainties, environmental disturbances, communication constraints, and collision avoidance (avoiding static and dynamic obstacles while maintaining the predefined formation pattern) \cite{peng2020overview}.

\begin{table*}[]
\caption{The Comparison of DL-Based or Related Cooperative Control Methods}
\resizebox{\textwidth}{!}{
\label{nn_coop_controllers}
\begin{tabular}{|l|l|l|l|l|l|l|l|}
\hline
\textbf{App. of DL} & \textbf{Controller} & \textbf{Proposed Model} & \textbf{ARC.} & \textbf{DIS.} & \textbf{CON.} & \textbf{ACT.} &  \textbf{REF.}                                   \\ \hline
\multicolumn{8}{|l|}{\textit{\textbf{Leader-follower formation control - Path-guided coordinated controls}}}                                                                               \\ \hline
~~MLP / E. &  Backstepping& NN-based robust adaptive formation controller&Cent.&\checkmark&-&Un.& \cite{peng2011robust}\\ \hline
~~MLP / E. &  DSC& Adaptive dynamic surface control&Cent.&\checkmark&-&Un.& \cite{peng2012adaptive}\\ \hline

\multicolumn{8}{|l|}{\textit{\textbf{Leader-follower formation control - Trajectory-guided coordinated controls}}}    \\ \hline

~~RBF / E.  &Adaptive robust control& Leader-follower formation tracking controller&Cent. &\checkmark&In.&Un.& \cite{shojaei2015leader} \\ \hline
~~MLP / E.  &Adaptive robust control& Output feedback formation control &Cent.&\checkmark&In.&Act.& \cite{shojaei2016observer} \\ \hline
~~RBF / E. &DSC&Robust adaptive formation control &Cent.&\checkmark&-& Un.&\cite{lu2018robust} \\ \hline
~~RBF / E. &DSC \& Backstepping& Adaptive platoon formation control &Dece.&\checkmark&Per.CA&Act.& \cite{dai2017platoon}\\ \hline
~~RBF / E. &DSC \& Backstepping& Decentralized adaptive formation control &Dece.&\checkmark&Per.CA& Act.&\cite{he2018leader} \\ \hline
~~MLP / E. &DSC \& Backstepping& Output-feedback formation tacking control&Dece.&-&Com.CA& Act.&\cite{dong2020neural} \\ \hline
~~RBF / E. &Backstepping& Adaptive finite-time event-triggered control&Cent.&\checkmark&Out.Per.& Act.&\cite{fu2021adaptive} \\ \hline
~~RBF / E. &Fault-tolerant control& Neural finite-time formation control&Cent.&\checkmark&-& Un.&\cite{huang2021adaptive} \\ \hline

\multicolumn{8}{|l|}{\textit{\textbf{Leaderless formation control - Path-guided coordinated controls}}}           \\ \hline
~~MLP / E.&DSC& Cooperative path following controllers &Dist.&\checkmark&In.&-& \cite{wang2014adaptive} \\ \hline
~~MLP / E.  &DSC& NN based adaptive dynamic surface control &Dece.&\checkmark&Com.&-& \cite{wang2014neural} \\ \hline
~~MLP / E. &Backstepping& Adaptive bounded neural network controller &Dist.&\checkmark&In.&Un.&  \cite{gu2020adaptive}\\ \hline
~~MLP / E. & Model-free control&Integrated distributed guidance and learning control&Dist.&\checkmark&-& Un.&\cite{liu2021distributed} \\ \hline
~~DRL / G. &DRL&USV Formation and Path-Following Control&Dist.&-&-&Un.&\cite{zhao2021usv} \\ \hline

\multicolumn{8}{|l|}{\textit{\textbf{Leaderless formation control - Trajectory-guided coordinated controls}}}           \\ \hline

~~MLP / E. &Backstepping& Distributed adaptive controller &Dist.&\checkmark&-&-&\cite{peng2013leaderless} \\ \hline
~~RBF / E. &DSC& Distributed robust adaptive cooperative control&Dist. &\checkmark&In.&-&\cite{lu2018adaptive} \\ \hline
~~MLP / E. &Backstepping& Distributed cooperation formation control &Dist.&\checkmark&Per.& Act.&\cite{liu2020cooperative} \\ \hline
~~WNN / E. &Backstepping& Distributed coordinated tracking control &Dist.&\checkmark&CA& Un.&\cite{liang2020distributed} \\ \hline
~~MLP / E. &DSC \& Backstepping& Adaptive neural formation control&Dece.&-&CA& Un.&\cite{he2021adaptive} \\ \hline
~~MLP / E. &Distributed coordinated control&Event-triggered distributed coordinated control&Dist.&\checkmark&In.& Act.&\cite{zhang2021event} \\ \hline
~~MLP / E. &Event-triggered control& Event-triggered adaptive neural fault-tolerant control&Cent.&\checkmark&In.& Un.&\cite{zhu2021event} \\ \hline

~~RBF / E. &SMC& Finite-time distributed formation control&Dist.&\checkmark&In.& Act.&\cite{huang2021finite} \\ \hline


\multicolumn{8}{|l|}{\textit{\textbf{Leaderless formation control - Maneuvering-guided coordinated controls}}}           \\ \hline
~~RNN / E.  &DSC& Containment maneuvering controller &Dist.&\checkmark&-&Act.& \cite{peng2016containment}\\ \hline
~~RNN / E.  &TD& Cooperative path maneuvering controller &Dist.&\checkmark&-&Un.& \cite{liu2017modular} \\ \hline
~~RNN / O. &Fuzzy kinetic control& Distributed maneuvering controller &Dist.&\checkmark&Vel.&-&  \cite{peng2017distributed}\\ \hline
~~RBF / E. &TD& Event-triggered modular-ISS NN controller &Dist.&\checkmark&-&Act.&  \cite{zhang2020event}\\ \hline
~~RNN / O. &Distributed control& Safety-critical containment maneuvering control&Dist.&\checkmark&In.CA& Un.&\cite{gu2021safety} \\ \hline

\end{tabular}
}

\footnotesize{Notes: 1. (-) not mentioned; (\checkmark) considered. 2. (App. of DL) The applied DL model and the applications of DL model, the applications include Model Uncertainties Estimation [E.], Control Signals Generation [G.], Optimization [O.]; (ARC.) Architecture in coordinated control, including Centralized control [Cent.], Decentralized control [Dece.], and Distributed control [Dist.]; (DIS.) Disturbance; (CON.) Constraints, including constraints on Input control signal [In.], Output control signal [Out.], Velocities [Vel.], Performance [Per.], Communications [Com.], Collision Avoidance [CA]; (ACT.) Actuated [Act.] or Underactuated [Un.]; (REF.) References.}
\end{table*}

\subsubsection{Leader-Follower Formation Control} 


Path-guided coordinated controls \cite{peng2011robust,peng2012adaptive} and trajectory-guided coordinated controls \cite{shojaei2015leader,shojaei2016observer,dai2017platoon,he2018leader,lu2018robust,dong2020neural,fu2021adaptive,huang2021adaptive} are two basic tasks for DL-based leader-follower formation control. The DL model is generally applied to estimate the uncertainties of the model, i.e., to compensate for unknown disturbance \cite{shojaei2015leader,shojaei2016observer,lu2018robust,fu2021adaptive}, unknown dynamic \cite{shojaei2015leader,peng2012adaptive,dai2017platoon,he2018leader,dong2020neural,fu2021adaptive,huang2021adaptive}, or unknown velocities \cite{peng2011robust,peng2012adaptive}.


\textbf{Path-guided Coordinated Controls:} 
Existing research attempted to address unknown dynamics, unmodeled disturbances, velocity estimations, and system stability in path-guided coordinated control tasks.
Peng \textit{et al.} \cite{peng2011robust,peng2012adaptive} proposed a DNN-based formation control that only uses the LOS range and angle measured by local sensors and can compensate for both uncertain leader and local dynamics.

\textbf{Trajectory-guided Coordinated Controls:} This research mainly focuses on the problems of unmodeled dynamics \cite{shojaei2015leader,shojaei2016observer,dai2017platoon,he2018leader,dong2020neural,fu2021adaptive,huang2021adaptive}, unknown disturbance \cite{shojaei2015leader,shojaei2016observer,dai2017platoon,he2018leader,fu2021adaptive}, actuator saturation \cite{shojaei2015leader,shojaei2016observer}, system stability \cite{shojaei2015leader,shojaei2016observer,dai2017platoon,dong2020neural,huang2021adaptive}, velocity estimation \cite{shojaei2016observer,huang2021adaptive}, collision avoidance \cite{dai2017platoon,he2018leader,dong2020neural}, connectivity maintenance \cite{dai2017platoon,dong2020neural}, performance constraints \cite{dai2017platoon,he2018leader}, computational effort reduction \cite{lu2018robust}, communication reduction \cite{dong2020neural,fu2021adaptive}, and output constraints \cite{fu2021adaptive}.


Taking into account the collision constraints, Dai \textit{et al.} \cite{dai2017platoon} limited the ASV position outputs to a given range and applied the prescribed performance control to guarantee that formation errors remained always within the predefined regions. Based on the DSC technique, the backstepping procedure and Lyapunov synthesis, adaptive formation control integrates DNN and disturbance observers to estimate unknown dynamics and disturbances, respectively.


To reduce communication cost, Dong \textit{et al.} \cite{dong2020neural} considered a one-to-one communication topology with a decentralized adaptive output feedback formation tracking controller, where DNN was used to approximate uncertain dynamics. 

\subsubsection{Leaderless Formation Control} 
The leaderless formation refers to a situation in which all agents reach a common value through local interaction with desired relative deviations and there is no physical leader among agents \cite{peng2013leaderless}. Therefore, the single point failure problem, which occurs when a leader does not work and the entire fleet cannot maintain a formation, can be avoided.
There are three basic tasks for DL-based leaderless formation control in this part, i.e., path-guided \cite{wang2014neural,gu2020adaptive,wang2014adaptive,liu2021distributed,zhao2021usv}, trajectory-guided \cite{peng2013leaderless,lu2018adaptive,liu2020cooperative,liang2020distributed,he2021adaptive,zhu2021event,huang2021finite,zhang2021event}, and maneuvering-guided coordinated controls \cite{peng2016containment,liu2017modular,peng2017distributed,zhang2020event,gu2021safety}. The DL model is generally applied to estimate the uncertainties of the model, that is, to compensate for unknown disturbance \cite{wang2014adaptive,wang2014neural,peng2016containment,liu2017modular,liang2020distributed,huang2021finite}, unknown dynamic \cite{peng2013leaderless,wang2014adaptive,wang2014neural,peng2016containment,liu2017modular,lu2018adaptive,liu2020cooperative,liang2020distributed,zhang2020event,he2021adaptive,zhu2021event,liu2021distributed,zhang2021event}, unknown input coefficients \cite{zhang2021event}, to solve the quadratic optimization problem \cite{peng2017distributed,gu2021safety} or to generate the control signal \cite{zhao2021usv}.

\textbf{Path-guided Coordinated Controls:} Based on different assumptions about what information is known or unknown by vehicles, existing research has made efforts to address unknown dynamics \cite{wang2014adaptive,wang2014neural,gu2020adaptive}, unmodeled disturbances \cite{wang2014adaptive,zhang2021event,wang2014neural,gu2020adaptive}, unknown kinetic models \cite{liu2021distributed}, input saturation \cite{wang2014adaptive,gu2020adaptive}, velocity estimation \cite{wang2014neural,liu2021distributed}, communication reduction \cite{wang2014adaptive,wang2014neural}, cyber attack \cite{gu2020adaptive}, system stability \cite{wang2014adaptive,wang2014neural,gu2020adaptive}, and control signal generation \cite{zhao2021usv}. 

With partial knowledge of the reference velocity, Wang \textit{et al.} \cite{wang2014adaptive} designed a cooperative path following control, which used the DNN-based DSC technique to estimate unknown dynamics and disturbances, used an auxiliary design to handle input saturation, and applied a distributed speed estimator to reduce the amount of communication. Cyberattacks exist widely in real-world communication networks. To achieve a desired formation during a state-dependent cyberattack varying in time, Gu \textit{et al.} \cite{gu2020adaptive} developed a path update law based on a synchronization scheme and an adaptive control method. DNN was used to approximate the uncertainties of the model and environmental disturbances.

The DRL model is able to resolve the problem of following the ASV formation path by designing reward functions that consider the velocity and error distance of each ASV related to the given formation \cite{zhao2021usv}. ASVs are able to automatically and flexibly adjust their formation under the proposed DRL-based method.

\textbf{Trajectory-guided Coordinated Controls:} Existing studies on trajectory-guided coordinated controls usually attempt to solve the problem of unmodeled dynamics \cite{lu2018adaptive,liang2020distributed,liu2020cooperative,he2021adaptive,zhu2021event,huang2021finite,zhang2021event}, unknown disturbance \cite{lu2018adaptive,liang2020distributed,liu2020cooperative,zhu2021event,huang2021finite,zhang2021event}, actuator saturation \cite{lu2018adaptive,zhu2021event,huang2021finite}, system stability \cite{peng2013leaderless,lu2018adaptive,liang2020distributed,liu2020cooperative,he2021adaptive,zhu2021event,huang2021finite,zhang2021event}, velocity estimation \cite{he2021adaptive}, collision avoidance \cite{liang2020distributed,he2021adaptive}, connectivity maintenance \cite{he2021adaptive}, unknown input coefficients \cite{zhang2021event}, computational effort reduction \cite{lu2018adaptive,liu2020cooperative,zhu2021event,huang2021finite}, self-organized aggregation \cite{liang2020distributed}, and communication reduction \cite{zhang2021event}.

Furthermore, to reduce computational complexity, researchers tried to decrease the number of learning parameters of DNNs with the minimum learning parameter algorithm \cite{lu2018adaptive,huang2021finite}, self-structured NNs \cite{liu2020cooperative}, or the virtual parameter leaning algorithm \cite{zhu2021event}. To save system resources, periodic communication-based event-triggered mechanisms were proposed in \cite{zhang2021event}. The desired path was predicted by the last event-triggered velocity during the triggering interval, and the model uncertainties and unknown input coefficients were estimated by a neural predictor based on concurrent learning. 

\textbf{Maneuvering-guided Coordinated Controls:} Existing studies tried to solve the following problems in maneuvering-guided coordinated control scenarios: Unmodeled dynamics \cite{peng2016containment,liu2017modular, peng2017distributed,gu2021safety,zhang2020event}, environmental disturbances \cite{peng2016containment,liu2017modular,peng2017distributed,gu2021safety,zhang2020event}, system stability \cite{peng2016containment,liu2017modular, peng2017distributed,zhang2020event,gu2021safety}, 
collision avoidance \cite{gu2021safety},
input saturation\cite{gu2021safety}, and communication reduction \cite{zhang2020event}. 


ASV has only limited resource, therefore, it is very practical to design a resource-constrained system. Instead of periodically updating the communication and actuation of systems, Zhang \textit{et al.} \cite{zhang2020event} designed an event-triggered DNN controller that decreased the communication burden of both followers and leaders. Under distributed directed communication, the actuator of each follower was updated when predetermined events are triggered. They utilized DNN to identify uncertain nonlinearities and introduced third-order linear tracking differentiators to estimate derivative information of the virtual control law. 

Taking into account the collision constraints, Gu \textit{et al.} \cite{gu2021safety} addressed the avoidance of collisions between vehicles and obstacles with input-to-state safe control barrier functions that mapped the safety constraints in states to the constraints in the control inputs. Furthermore, to calculate the forces and moments, the quadratic optimization problem was solved using an RNN-based neurodynamic optimization approach.

\subsubsection{Data-driven IoT System}
Modern industrial systems with various IoTs generate rich maritime data that should be appropriately analyzed to improve both the efficiency and reliability of existing systems. Perera \textit{et al.} \cite{perera2020deep} designed a general framework for autonomous ship navigation for ASVs to achieve the required level of ocean autonomy. Each on-board ASV application may be equipped with an on-board decision-making process and is monitored by on-board and onshore IoT under a DL-type framework.

\subsubsection{Maritime Traffic Monitoring and Prediction}
Forecasting future traffic in a complex maritime environment can help design ship routes, reduce traffic jams, and improve the efficiency of traffic management, especially for inland waterways. Surface vehicles have different dimensions, shapes, and boundaries in physical form and can travel in a two-dimensional spatial surface, making maritime traffic monitoring and prediction more difficult \cite{xiao2019traffic}. The given maritime region can be divided into grids, and then the inflow and outflow of each grid can be predicted. To predict the inflow and outflow of all grids, Zhou \textit{et al.} \cite{zhou2020using} studied the spatial and temporal dependencies of the vessel flows by extracting the spatial features of the patterns of maritime traffic with CNN and learning the temporal correlation of the extracted patterns using LSTM. 

\subsection{Communication and Networks}
The ecosystem elements in cooperative operations include ASVs, UAVs, and AUVs. Advanced equipment is capable of collecting large amounts of data volume in different forms, such as images, videos, audio, and text, which introduces a great burden for the maritime service. Yang \textit{et al.} \cite{yang2019novel} proposed a software-defined networking (SDN)-based framework that applies DRL to solve the overfitting and curse of dimensionality by establishing a mapping relationship between the acquired information and the optimal data transmission scheduling strategy in a self-learning way.

\subsection{Energy Efficient Operations}
Reducing the speed of ships is the most effective way to improve energy efficiency. However, the best real-time speed of the ship is related to several factors, including future environmental conditions. In \cite{yuan2021prediction}, data from AIS, GPS, and fuel, rotation, temperature, and environment sensors were obtained by continuous time sampling. Then, on the basis of LSTM, these data were used to calculate the fuel consumption rate of ships in real time. In addition, an optimization algorithm was presented, which is called the reduced space search algorithm, to minimize fuel consumption and the total cost of a voyage.

\section{Current Challenge and Future Perspectives}
\label{sec:challenge}
This section presents current challenges and potential future directions for NGC systems, cooperative operation, application scenarios, and DL limitations. 

\subsection{ASV Prototypes and Their Applications}

We comprehensively compare the NGC system, communication method, mechanical design and applications of 51 ASV prototypes, which can be found on the author's website\footnotemark[1]. 
By reviewing typical ASV prototypes and comparing their key features, we can see the driving force of ASVs.
\footnotetext[1]{The comparison of commercial and research ASV projects, https://yuanyuanqiao.github.io/publications/journal/qiao2022survey\_appendix.pdf}

\subsection{NGC system}

\subsubsection{Navigation}
DL-based methods have dramatically improved state-of-the-art perception in recent years \cite{lecun2015deep}. However, perception remains very challenging due to harsh maritime environments.



\textbf{Environmental Perception:} For safer maritime navigation, ASVs must be aware of the surrounding environment in
real time. An open dataset with large amounts of labeled data for marine navigation is not yet available. Therefore, most research on DL-based marine environmental perception focuses on remote sensing images that have open and labeled datasets. Image preprocessing, segmentation, and target recognition methods are the main topics of remote sensing processes. 
DL methods can extract and fuse low- and high-level features from raw datasets to find the complex relationship between the data by self-learning. Based on the growing interest in object detection on SAR images in the last 3 years, DL-based methods have been explored in depth to solve the inherent problems of remote sensing images, such as detecting dense and overlapping objects, small and densely clustered ships, and complex backgrounds \cite{ball2017comprehensive}. 
However, to detect objects with complex backgrounds and multiple shapes in marine environments, learning robust and discriminative representations from complex remote sensing images is still very challenging compared to natural scene images\cite{zhang2016deep}.

In addition, for the data collected from onboard sensors, such as optical, infrared, radar images, and point clouds data, existing methods regarding image classification, object detection, and segmentation can be well adapted to maritime offshore environment, which is a much more simple scenario in contrast to complex road conditions for autonomous cars. In marine environments, the vibration interference generated by non-stationary surface platforms (buoys, sailing ships, etc.) can be easily removed. 
DL techniques have been shown to outperform other conventional methods in accurate and robust detection \cite{yao2019lidar}, and performance can be improved with more labeled data \cite{han2020autonomous}. DL-based methods are expected to be adapted to complex computer vision problems for navigation in marine environments, such as vision-based multisensor fusion for cooperative operations and berthing.

\textbf{State Estimation:} Estimating the states of focal ASV and other ASVs requires a data fusion technique, which is very important for path following, trajectory tracking, and multi-vehicle cooperation. Existing research mainly focuses on estimating the state of ASV with time series data, such as trajectory and motion data collected from AIS. Very few researchers try to fuse maritime video and AIS data to better understand the on-site traffic situation awareness information  \cite{chen2020video}. Due to the strong capability in feature extraction and data representation, DL-based methods have already achieved high accuracy on data fusion tasks, especially for complex and imprecise data 
and image.
How to extract and fuse features of maritime multi-sensor data with DL methods is very challenging, especially when the collected data is discontinuous and incomplete, which can be solved by the application of CNN, LSTM, AE, attention, and other DL models.
Another possible research direction is to detect out-of-distribution (OOD) sensor data before fusion to achieve a better results. Furthermore, understanding the semantic environment, which is receiving increasing attention in the robotic field, can improve the efficiency of the calculation. 
DL has made great progress in image understanding 
and semantic segmentation 
and it is anticipated that they will be applied to maritime navigation to reduce the reliance on complex multisource data \cite{zhou2020review}.

Future development of environmental perception and state estimation for ASV navigation can still benefit greatly from the existing results of other autonomous vehicles. Based on transfer learning \cite{zhuang2020comprehensive}, the knowledge of other autonomous vehicles can be utilized to navigate ASVs to further improve the integrity level of ocean autonomy. In this article \cite{perera2020deep}, the DL-type mathematical framework is considered the best tool for mimicking helmsman actions in ship navigation. However, how corresponding decisions are made to support each ASV's different navigation situations is an open question.

\subsubsection{Guidance}

Traditional path planning methods can enumerate appropriate trajectories between two points for ASV if all constraints, such as geographic features, static obstacles, dynamics and kinematics of ASV and energy consumption, have already been detected and considered. For complex tasks with uncertain and incomplete information in constantly changing environments, heuristic methods such as DNN can achieve optimal solutions. 
For global path planning, DL-based methods are able to learn patterns from massive historical data \cite{xiao2019traffic}. In addition, DRL can learn to react by interacting with an unknown environment to find the most efficient route for local path planning. 
The DRL network is able to evaluate possible behaviors and make the next move in a collision avoidance scenario. 
Furthermore, DL-based methods are very promising for addressing collision avoidance by analyzing complex and ambiguous situations when ASVs encounter other ships \cite{woo2020collision} and can learn behavior patterns from historical data \cite{cheng2018concise}. The challenges in terms of DL application in guidance systems are discussed below.



\textbf{COLREGs Compliance:} The risk assessment and solution for COLREG-compliant collision avoidance in multiple-ship encounter situations cannot be solved by conventional methods if unknown disturbances and uncertain ship motions are impossible to ignore \cite{woo2020collision}. Most current experiments focus on simple encounter situations that only consider the OS and one TS based on Rules 6, 8, 13-19 of COLREGs. DL-based methods have already made some promising progress in collision avoidance, especially for situations that need to comply with COLREGs, which require human-like decision-making \cite{zhao2019colregs,shen2019automatic,zhang2021collision}, because they are capable of extracting high-level characteristics from complex environments and learning the relationship between situations and possible actions. However, the following problems are very challenging. 

\begin{itemize}

\item \textbf{The design of reward functions.} The architecture of a DRL network requires careful design and is usually based on experience and practice, rather than mathematical reasoning, resulting in inflexible application in real systems.

\item \textbf{No universal rule for collision avoidance.} The highly complex dynamics under different environmental conditions vary with the types and sizes of ships that require specific rules for each encounter. 

\end{itemize}

As a result, apart from proposing a model that makes human-like decisions to comply with COLREGs, designing maritime traffic rules that adapt to modern maritime environments has become a more reachable goal in recent years. Then, the DL model can be used to predict maritime traffic and the trajectories of sailing ASVs, which is important for maritime traffic management.

\textbf{Decision Making:}
Real-time decision-making refers to the ability to make appropriate decisions under practical maritime conditions based on knowledge of the surrounding environment of the ASV. In collision avoidance \cite{woo2020collision}, which has challenges including uncertainties in ship motion models, rule-compliant navigation systems, and unknown environmental disturbances \cite{huang2020ship}, it is very difficult to evaluate all possibilities and find the optimal behaviors based on the massive data collected by multiple sensors. On the one hand, nonlinear methods are often computationally expensive, require \textit{a priori} knowledge of unknown states, and are driven by human knowledge \cite{everett2018motion}. On the other hand, DL methods have already made many processes in decision-making for autonomous vehicles.
In the future, DL-based integrated approaches for robots are expected to reduce uncertainty in perception, decision-making, and execution \cite{sunderhauf2018limits}.


\textbf{Cooperative Path Planning for Multiple ASVs:} Multiple ASVs can perform more complex tasks than a single ASV, requiring the analysis of emergent behaviors, learning communication, learning cooperation, and agent modeling. 
Conventional cooperative methods are usually limited to discrete actions and require handcrafted features, which might not be suitable for real applications. 
On the other hand, DRL methods can generate paths for multiple ASVs to keep the formation shape robust or vary the shapes where necessary.

\subsubsection{Control}

Unlike cars or other modes of land transportation, one of the biggest problems for ASV is the hydrodynamic water-body interaction. Although this interaction complicates the problems of motion and control, DL-based controllers are introduced as a potential solution. In general, DNNs are required to estimate unknown parameters, terms, or functions during ASV modeling. In addition, other approximators, such as fuzzy systems, can also obtain similar results to DNNs. However, the exact mathematical estimator may not be able to adapt to complex sea environments \cite{zhao2020path}. 
For the DRL-based model, the agent interacts with the unknown or potentially partially unobservable environment and then iteratively approximates the behavior policy that maximizes the agent’s expected long-term reward in the environment. 
Although it has gained much attention in the field of autonomous systems \cite{kiran2021deep}, the DRL-based controller cannot yet support real-world ASV. 
Most of the above work tests DRL-based control in simulation environments, which provides a large-scale training dataset. However, machine learning models usually perform poorly if the data distributions of the training dataset and testing dataset are not identically distributed; thus, ASV training in simulation environments often fails to complete tasks in the real world. A possible solution is to use transfer learning, which applies an algorithm trained in one or more ``source domains" to a different (but related) ``target domain" \cite{zhuang2020comprehensive}. Knowledge learned from other vehicles or other tasks can be used by employing transfer learning, which requires fewer training data for the current task. Therefore, pre-training agents on available high-quality datasets and then fine-tuning them on collected datasets can improve the performance of the model. 
However, due to the lack of open datasets, the related research field is still in its early stages.

Although RL-based controllers have been successfully applied to ASVs on sea trails, 
DRL-based controllers are still in an early stage and can only be used to estimate the parameters of a model that presents a physical system. More complex problems, such as such constraints, have not yet been considered. To construct a complete real-world ASV system, better performance can be achieved by combining DL and classical controllers such as MPC or PID \cite{grigorescu2020survey}. All hard constraints of the modeled system are estimated by DL, and classical model-based control techniques provide a stable and deterministic mode \cite{woo2020collision}.

\subsection{Cooperative Operations}

A fleet of ASVs is able to complete more difficult tasks than a single ASV. However, the design of the controller and communication scheme for cooperative control of multiple ASVs is very challenging, since ASVs must maintain desired positions and orientations with predefined geometric shapes \cite{peng2020overview}. Since the DL-based model has already been applied to design collision-free cooperative controllers and communication protocols for multi-agents \cite{everett2018motion}, improving these methods for maritime environments is anticipated in the future. The challenges of cooperative operations and the potential of DL applications are discussed
as follows.

\subsubsection{Behavior Prediction}
Massive information collected by AIS provides an opportunity to discover the historical behavior of fleets for the management of maritime traffic safety \cite{xiao2019traffic}, trajectory reconstruction, anomaly detection and vessel type identification. In addition to a large amount of multi-source data stream, the inherent noise patterns and irregular time sampling, and the influence of weather and current have also increased the difficulty of analyzing AIS data. DL methods have already been applied to road traffic prediction because they can capture the spatial and temporal correlations of traffic
while simultaneously considering the influences \cite{koesdwiady2016improving}. Therefore, how to make the best use of the increasing amount of AIS data has great significance for maritime cooperative operations, which requires the assistance of advanced methods such as DL.

\subsubsection{Communications and Networks}
\begin{figure}
\centering 
\includegraphics[width=0.35\textwidth]{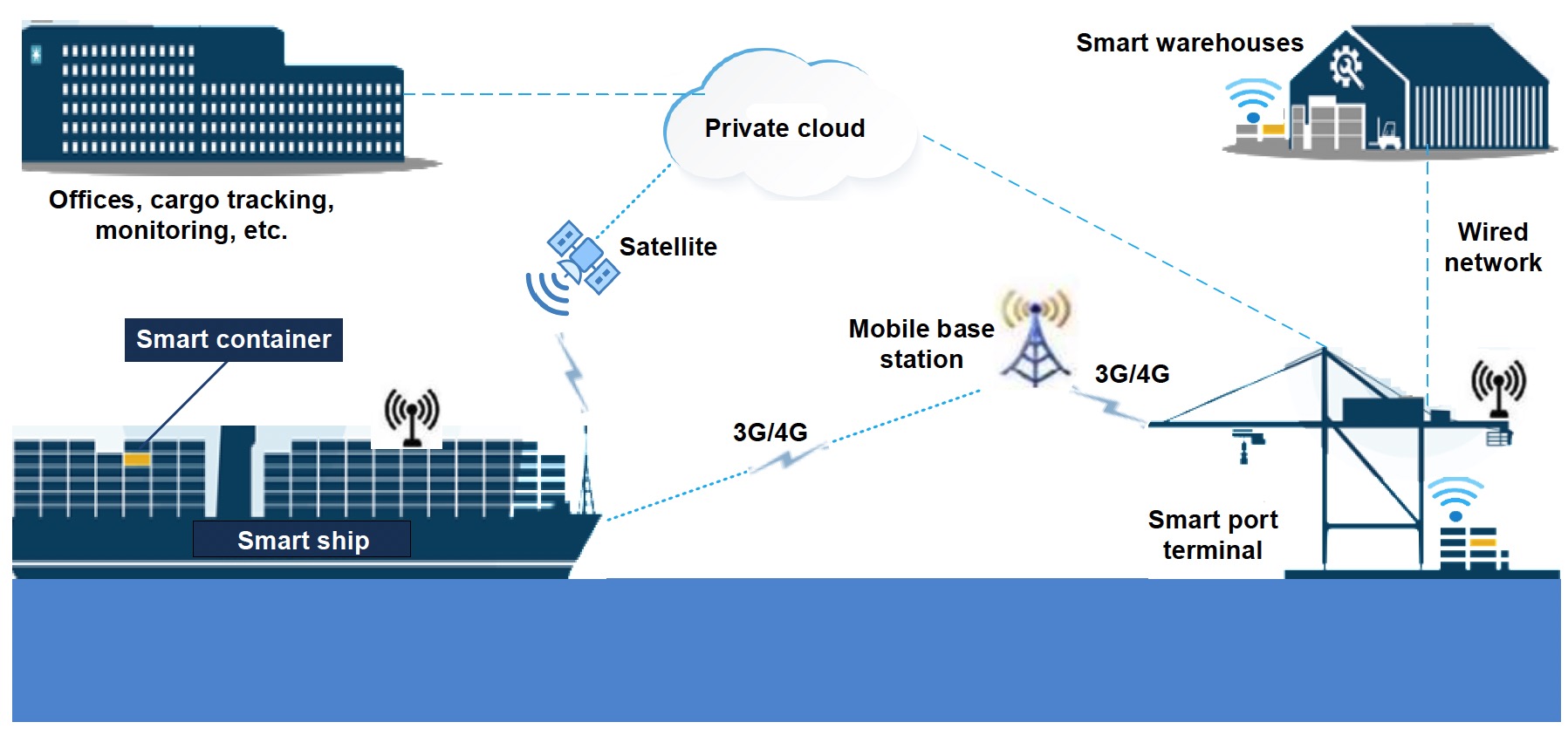} 
\caption{An example of Internet of Ship scenario \cite{aslam2020internet}} 
\label{ios} 
\end{figure}

GPS has provided a high-performance navigational aid, and commercially available satellite communication receivers allow for almost instantaneous communication over most of the world. Networking of ASVs and UAVs further expands the communication range on the surface \cite{zolich2019survey}. Furthermore, the network of smart interconnected maritime objects has received a new concept term, the Internet of Ships (IoS) \cite{aslam2020internet}, as shown in \figurename~\ref{ios}. The main technical challenge for remote operation is the connectivity between the vessel and the control facility.
Possible solutions are to build high-bandwidth and low-latency wireless networks sufficiently and to reduce large-scale sensor data. DL-based methods have already been used to design wireless networks, 
compress data, and determine an optimal strategy for data transmission \cite{yang2019novel}. These methods are expected to be applied to future maritime network management and control.

\subsection{Specific Application Scenarios}

\subsubsection{Smart Port}

How ASVs will berth and maneuver around densely trafficked ports has gradually become an imminent problem for modern maritime security and management. Existing work has already applied the latest technologies to smart ports \cite{yang2018internet}. There are many computer vision tasks in port, such as container identification, image and object recognition, situational awareness, monitoring, and surveillance \cite{aslam2020internet}. DL-based methods can help develop a fully automated port with advanced sensing systems. Future work related to ASV monitoring and management in ports, MASS docking and departure, and loading and unloading are anticipated.

\subsubsection{ASV in City}
ASV may also be able to play an important role in the future of transportation in many coastal and riverside cities, such as Amsterdam and Venice, where the existing infrastructure of roads and bridges is always extremely busy. The Roboat project \cite{wang2020roboat} will develop a logistics platform for people and goods, superimposing a dynamic infrastructure in one of the world’s most famous water cities. 
When ASVs move in narrow and crowded urban environments, the requirements of autonomous systems, such as control and obstacle avoidance, are much higher than those of surface vehicles for open-water areas. Therefore, a DL-based framework, especially DRL, could be a viable option for ASV control and obstacle avoidance problems in these challenging urban environments.

\subsubsection{MASS}
Related research is still in the conceptual stage and focuses mainly on discussing the operation and safety of MASS, including navigational risk \cite{fan2020framework}, collision avoidance \cite{ramos2019collision}, and human-system interaction in autonomy \cite{ramos2020human}. Most of the navigation algorithms deployed on small ASVs cannot be applied directly to MASS. Large ships need more sensors than small ASVs to cover the very large MASS hull. Furthermore, MASS deceleration at moderate or high speeds, during which time the capabilities of the thrusters are negligible, requires more time and resistance. This makes the design of the controller and collision avoidance methods for MASS entirely different from other vehicles. The above-mentioned problems may be overcome by mathematical frameworks based on DL supported by the decision support layer \cite{perera2020deep}. More specifically, communication and computation infrastructure is needed by adopting an information technology/information system (IT/IS) based on real-time data-driven decision support in the maritime industry. In addition, 
a large amount of ship performance and navigation data collected and exchanged by onboard and onshore IoT should be integrated with modern IT systems and DL algorithms. However, the limited availability of testing ships and fields has slowed the study of practical applications for MASS in the scientific research field.

\subsection{The Limitation of DL}
DL is considered one of the most promising methods for feature learning \cite{zhang2016deep}. Its limitations are as follows and possible solutions can be found in paper \cite{bengio2021deep}. 

\begin{itemize}

\item \textbf{Learning with little or no external supervision:} Supervised learning requires a lot of labeled data. DRL requires far too many experiments and handcrafted rewards.

\item \textbf{Coping with test examples that come from a
different distribution than the training examples:} The assumption of Independent Identically Distribution (I.I.D) is central to almost all machine learning algorithms. Thus, DL models cannot quickly adapt to data distribution changes
with very few examples.

\item \textbf{Solving problems like humans:} It is very difficult for DL algorithms to solve problems by using a deliberate sequence of steps.

\end{itemize}

There are several methods that can alleviate the problems of DL in the marine environment. For the first and second problems, in addition to calling for a high quality open dataset or fine-tune networks pre-trained on ImageNet \cite{deng2019learning}, transfer learning and GAN have already been applied to reuse high-level feature during training 
and to generate high-quality samples, 
respectively. However, this process is still quite challenging, because bias will inevitably be introduced due to the domain mismatch between different types of data. In recent years, as a self-supervised learning algorithm, contrastive learning has gained increasing attention. 
This algorithm can learn the general features of a dataset without labels by learning representations such that similar samples remain close to each other. It would be interesting to pretrain the maritime navigation module under the contrastive learning framework, which can support various downstream tasks.

Furthermore, based on DRL, the control law can be learned directly to compensate for uncertainties and disturbances \cite{zhang2021model}. The reward functions of DRL have a very large impact on the learned desired behavior. However, the reward function is usually hand-crafted, and a single reward function cannot adapt to complex systems in a dynamic environment. Existing studies designed extrinsic rewards considering the following path \cite{meyer2020taming,zhao2020path}, obstacle avoidance \cite{meyer2020taming,cheng2018concise}, velocity\cite{zhao2020path,cheng2018concise}, environmental disturbances \cite{zhao2020path}, distance \cite{cheng2018concise}, etc. A better solution for the agent is to learn some intrinsic reward functions by itself. For example, the agent can perform imitation learning by incorporating domain knowledge into RL with
reward shaping,
or observing its behavior with inverse RL. 

\section{Conclusion}
\label{sec:conclusion}

Full autonomy can be anticipated in the following decades for maritime scenarios. However, common interfaces, which heavily rely on advances in the underpinning technology, have not yet been agreed upon. This paper aims to bridge the research gap between DL applications and ASVs by providing a systematic review of the literature on the overlap between these two fields. The importance of this work is emphasized by comparing existing ASV-related surveys. Then, state-of-the-art DL models are presented, as well as their implementation on NGC systems and maritime cooperative operations classified by scenarios. In consideration of the above-related studies, we discuss current challenges and possible future research topics in the direction of intelligent maritime autonomous operations.

\section{Acknowledgment}
The authors would like to thank the AMS Amsterdam Institute for Advanced Metropolitan Solutions and the members of the MIT Senseable City Lab consortium. This work is supported by the Funds of the National Natural Science Foundation of China (No. 62272057).

\bibliographystyle{IEEEtran}
\bibliography{reference}

\begin{IEEEbiography}[{\includegraphics[width=1in,height=1.25in,clip,keepaspectratio]{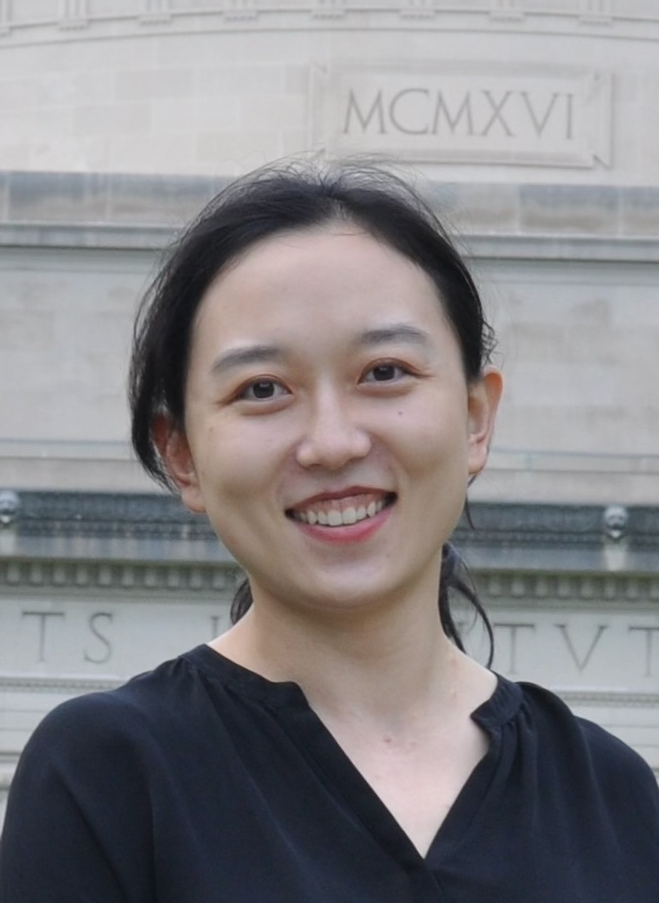}}]{Yuanyuan Qiao} 
(M'17)
received the B.E. degree from Xidian University, Xi’an, China, in 2009 and the Ph.D. degree from the Beijing University of Posts and Telecommunications (BUPT), Beijing, China, in 2014. From September 2019 to October 2020, she was a visiting scholar in Senseable City Lab, MIT, Cambridge, MA, USA.

She is currently an associate professor with the School of Artificial Intelligence, BUPT, Beijing, China. Her research focuses on deep learning based big data analytics.
\end{IEEEbiography}

\begin{IEEEbiography}[{\includegraphics[width=1in,height=1.25in,clip,keepaspectratio]{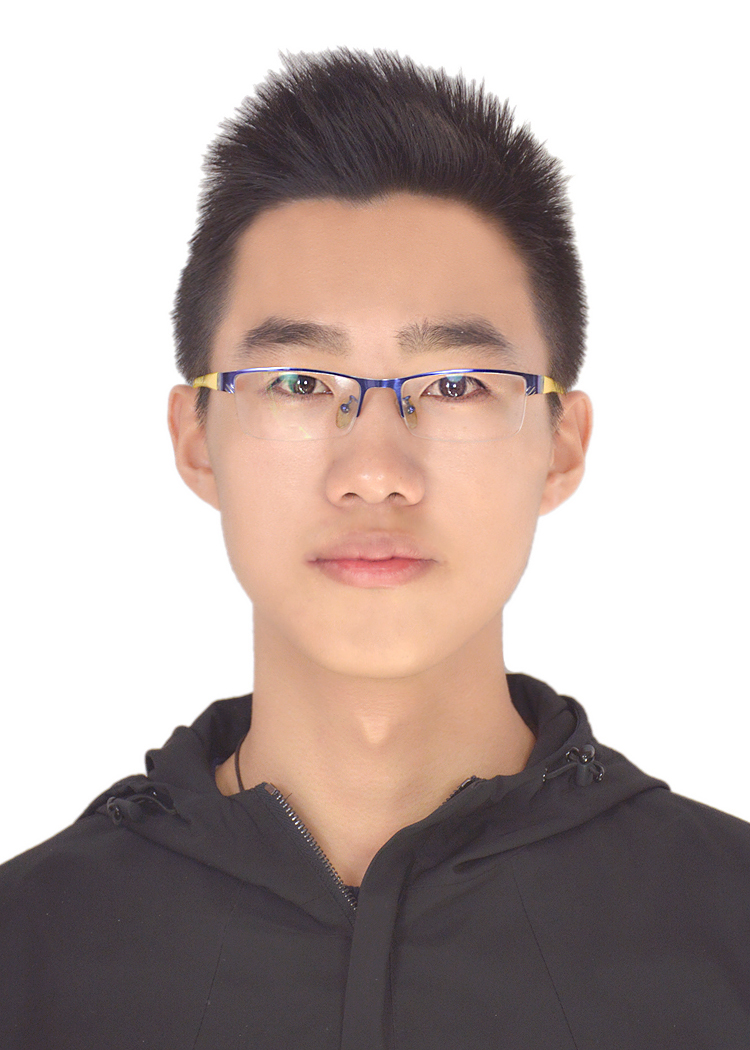}}]{Jiaxin Yin} received his Bachelor’s degree in School of Information and Telecommunication Engineering from BUPT, Beijing, China, in 2020.
 
He is currently a Ph.D. student from the School of Artificial Intelligence, BUPT, Beijing, China, since Sept. 2020, supervised by Prof. Jie Yang. His research topic contains semi-supervised anomaly detection, out-of-distribution detection and self-supervised learning.
\end{IEEEbiography}

\begin{IEEEbiography}[{\includegraphics[width=1in,height=1.25in,clip,keepaspectratio]{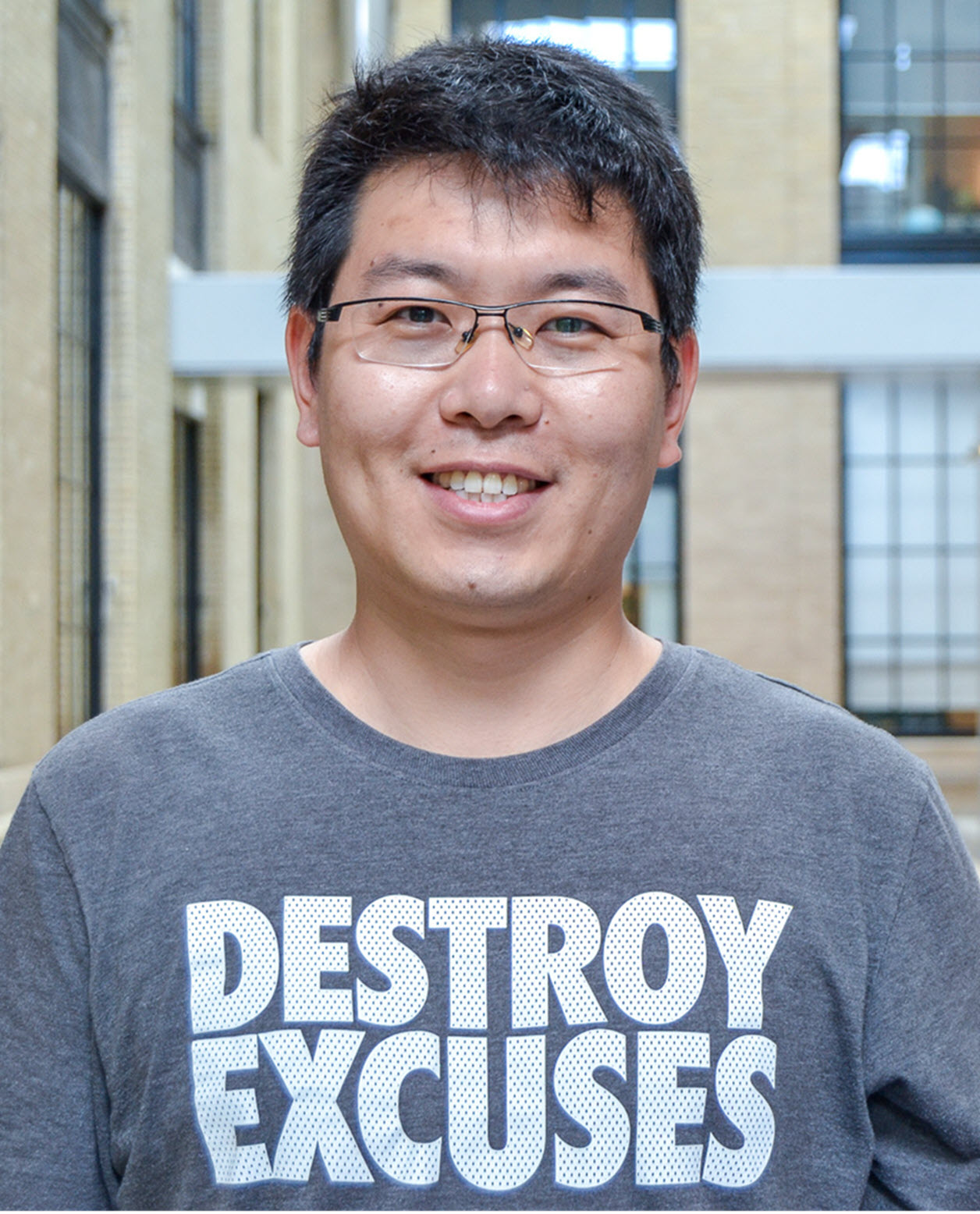}}]{Wei Wang} (M'13)
received the B.E. degree in automation from the University of Electronic Science and Technology of China, Chengdu, China, in 2010, and the Ph.D. degree in general mechanics and foundation of mechanics from Peking University, Beijing, China, in 2016. 

He was a joint Postdoctoral Associate from September 2016 to September 2019, and a Senior Postdoctoral Associate from September 2019 to September 2021 in both the Computer Science and Artificial Intelligence Laboratory (CSAIL) and the Senseable City Lab, MIT, Cambridge, MA, USA. He is currently a joint Research Scientist in both laboratories. His current research interests include marine robotics, dynamics and control, autonomous navigation, multi-robot systems, and bioinspired robotics.
\end{IEEEbiography}

\begin{IEEEbiography}[{\includegraphics[width=1in,height=1.25in,clip,keepaspectratio]{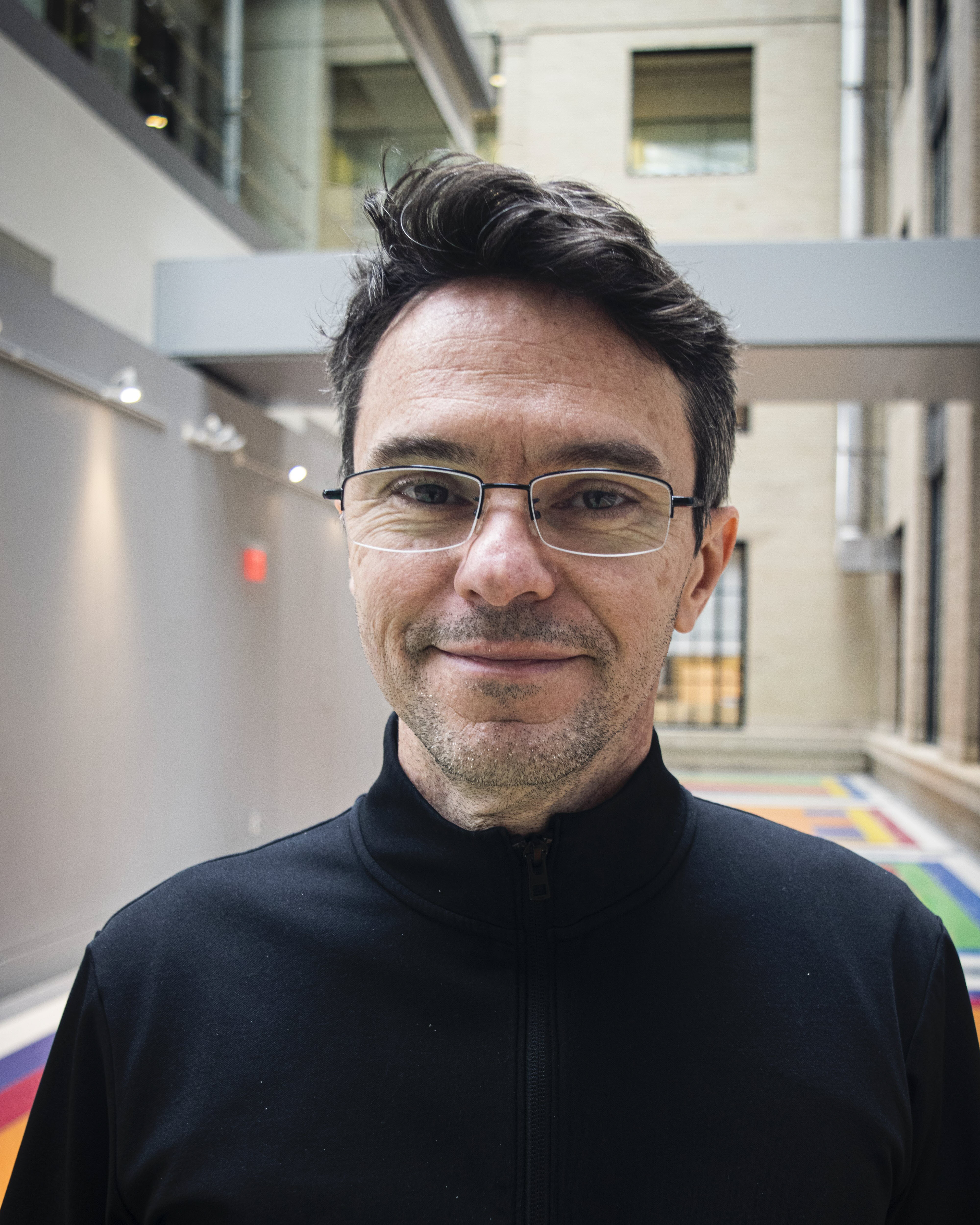}}]{Fábio Duarte} received the Ph.D. degree in communication and technology from the Universidade de São Paulo, Brazil. 
He has a background in urban planning and he is a Lecturer in DUSP, Head of Research Initiatives in the Center for Real Estate, and Principal Research Scientist at the MIT Senseable City Lab, where he manages projects including Underworlds, Roboat, City Scanner, as well as the data visualization team. Duarte was a professor at PUCPR (Curitiba, Brazil), and has been a visiting professor at Yokohama University and Twente University. 

Duarte serves as a consultant in urban planning and mobility for the World Bank. His most recent books are “Urban play: make-believe, technology and space” (MIT Press, 2021) and “Unplugging the city: the urban phenomenon and its sociotechnical controversies” (Routledge, 2018).
\end{IEEEbiography}

\begin{IEEEbiography}[{\includegraphics[width=1in,height=1.25in,clip,keepaspectratio]{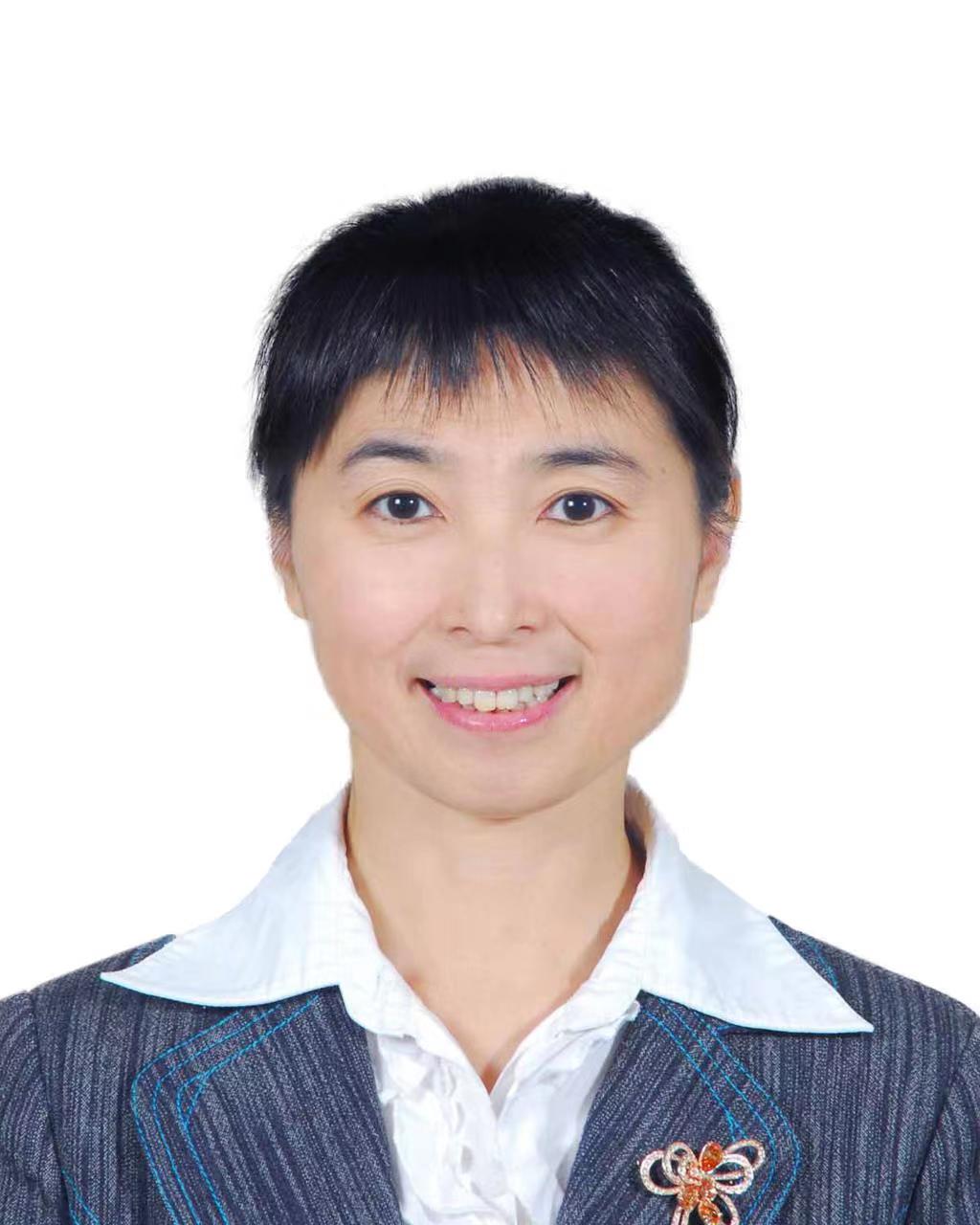}}]{Jie Yang} received the B.E., M.E., and Ph.D. degrees from the BUPT, Beijing, China in 1993, 1999, and 2007, respectively. 

She is currently a professor with the School of Artificial Intelligence, BUPT, and the director of the teaching and research center of intelligent perception and computing, BUPT. Her research focuses on deep learning based big data analytics.
\end{IEEEbiography}

\begin{IEEEbiography}[{\includegraphics[width=1in,height=1.25in,clip,keepaspectratio]{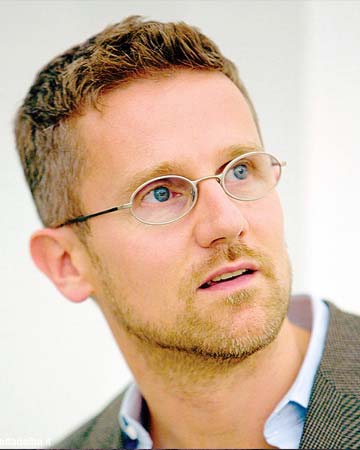}}]{Carlo Ratti} graduated from the Politecnico di Torino and the École nationale des ponts et chaussées, and received the M.Phil. and Ph.D. degrees from the University of Cambridge, U.K. He is currently the Founder and the Director of the MIT Senseable City Laboratory, an Architect, and an Engineer by training. He practices in Italy and teaches at the Massachusetts Institute of Technology.

He has coauthored over 200 publications and holds several patents. His work has been exhibited worldwide at venues, such as the Venice Biennale; the Design Museum Barcelona; the Science Museum, London; GAFTA, San Francisco; and The Museum of Modern Art, New York. His Digital Water Pavilion at the 2008 World Expo was hailed by Time Magazine as one of the “Best Inventions of the Year.” He has been included in Esquire Magazine’s “Best and Brightest” list, Blueprint Magazine’s “25 People Who Will Change the World of Design,” and Forbes Magazine’s “Names You Need To Know” in 2011. He was a Presenter at TED 2011. He is serving as a member for the World Economic Forum Global Agenda Council for Urban Management.

\end{IEEEbiography}

\end{document}